\documentclass[lettersize,journal]{IEEEtran}
\usepackage{amsmath,amsfonts}
\usepackage{algorithmic}
\usepackage{algorithm}
\usepackage{array}
\usepackage[caption=false,font=normalsize,labelfont=sf,textfont=sf]{subfig}
\usepackage{textcomp}
\usepackage{stfloats}
\usepackage{url}
\usepackage{verbatim}
\usepackage{graphicx}
\usepackage{cite}
\usepackage{soul}
\usepackage{xcolor}
\usepackage{longtable} 
\usepackage{supertabular}
\usepackage{rotating}

\usepackage{booktabs}
\usepackage{multirow}
\usepackage{graphicx}
\usepackage[normalem]{ulem}
\useunder{\uline}{\ul}{}
\usepackage{lscape}
\usepackage{adjustbox}
\usepackage{cleveref}

\usepackage{multirow}
\usepackage{array}
\newcolumntype{C}[1]{>{\centering\arraybackslash}p{#1}}

\begin{document}

\title{A Survey on Video Anomaly Detection via Deep Learning: Human, Vehicle, and Environment}

\author{Ghazal Alinezhad Noghre$^1$, Armin Danesh Pazho$^1$, Hamed Tabkhi$^1$
\thanks{$^1$ Electrical and Computer Engineering Department, UNC Charlotte (galinezh, adaneshp, htabkhiv@charlotte.edu)}
}

\markboth{IEEE Transactions On Pattern Analysis and Machine Intelligence}%
{Noghre \MakeLowercase{\textit{et al.}}: VAD via Deep Learning Survey}


\maketitle

\begin{abstract}
Video Anomaly Detection (VAD) has emerged as a pivotal task in computer vision, with broad relevance across multiple fields. Recent advances in deep learning have driven significant progress in this area, yet the field remains fragmented across domains and learning paradigms. This survey offers a comprehensive perspective on VAD, systematically organizing the literature across various supervision levels, as well as adaptive learning methods such as online, active, and continual learning. We examine the state of VAD across three major application categories: human-centric, vehicle-centric, and environment-centric scenarios, each with distinct challenges and design considerations. In doing so, we identify fundamental contributions and limitations of current methodologies. By consolidating insights from subfields, we aim to provide the community with a structured foundation for advancing both theoretical understanding and real-world applicability of VAD systems. This survey aims to support researchers by providing a useful reference, while also drawing attention to the broader set of open challenges in anomaly detection, including both fundamental research questions and practical obstacles to real-world deployment.
\end{abstract}


\bstctlcite{BSTcontrol}
\section{Introduction}
\label{sec:intro}

\IEEEPARstart{V}{ideo} Anomaly Detection (VAD), also known as outlier detection, abnormal event detection, and abnormal activity detection, has emerged as a crucial technology across a range of applications \cite{shaukat2021review, liu2024deep, mishra2021vt, yang2022visual, pazho2023survey}, from public safety \cite{rezaee2024survey, fahrmann2024anomaly, samaila2024video, mishra2024skeletal, pazho2023ancilia} to healthcare monitoring \cite{yang2023deep, ali2025anomaly, fernando2021deep, galvao2024anomaly}, autonomous driving \cite{bogdoll2022anomaly, baccari2024anomaly, bogdoll2023perception, solaas2024systematic}, road surveillance \cite{santhosh2020anomaly, rathee2023automated, khan2022anomaly}, and environmental disaster detection and response \cite{huang2022fire, saleh2024forest, gao2024two, liang2023v, filonenko2015real, lopez2018review}. In this age, where thousands of cameras continuously capture data, automated systems for detecting unusual events offer transformative potential \cite{ardabili2023understanding, ardabili2022understanding, pazho2024vt, ardabili2024exploring}. For example, surveillance VAD (see Supplementary Materials for list of abbreviations) can automatically flag crimes or accidents, relieving human operators of the impossible task of watching hours of mostly uneventful footage. Another example is in healthcare, where VAD can monitor patients or older adults for sudden falls or distress. The growing importance of VAD in such domains stems from its ability to consistently watch for anomalies that could signify security threats, medical emergencies, or catastrophic events.

VAD confronts unique challenges inherent to video data. A video is a high-dimensional spatiotemporal signal: each anomaly may involve not just an unusual appearance in a single frame, but an irregular motion pattern unfolding over time \cite{tian2021weakly,  wu2024vadclip, wang2021robust}. An anomaly in video can be formally defined as "The manifestation of atypical visual or motion characteristics, or the presence of typical visual or motion patterns occurring in a spatiotemporal contexts that deviate from established norms". An example of an abnormal pattern can be a car accident, which represents a deviation from expected vehicular operation. On the other hand, a normal pattern occurring in an inappropriate context is exemplified by riding a bicycle on a pedestrian-only sidewalk. Moreover, regardless of the specific domain or application area, anomalous events are inherently rare, often occurring with low frequency and unpredictability \cite{samaila2024video, ramachandra2020survey, yang2024follow, liu2018ano_pred, Rodrigues_2020_WACV, danesh2023chad}. VAD may encounter novel, unforeseen abnormal events that were never observed. Even new patterns of normal activity may continually emerge, especially in open-world environments \cite{zhu2022towards, alinezhad2023understanding, noghre2024exploratory, yao2024evaluating}. 

\begin{figure*}[t]
\centering
\includegraphics[clip,trim={18 18 18 18},width=0.85\textwidth]{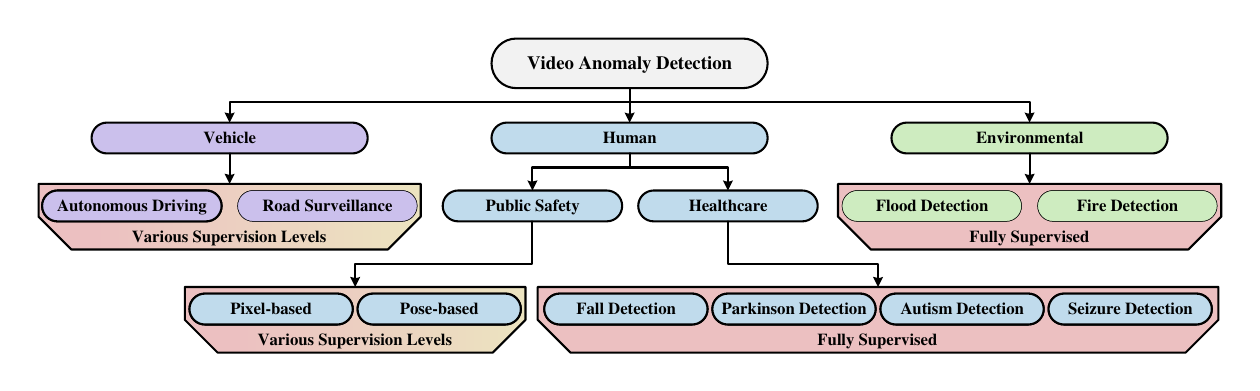}
\caption{Overview of the paper structure. The advancements in vehicle, human, and environmental VAD are explored.}
\label{fig:flow}
\end{figure*}

Traditionally, VAD relied on statistical models and handcrafted features to identify unusual patterns \cite{zhu2012context, zhou2007detecting}. These methods often struggled with the complexity and variability inherent in video data and previously mentioned general challenges, leading to limited accuracy and adaptability. Deep learning has been a driving force behind recent progress in VAD, enabling models to automatically learn rich representations of normal and abnormal patterns. A wide spectrum of learning paradigms, from fully supervised \cite{wang2020fall, gong2020novel, jin2020diagnosing, mehta2023privacy, kirichenko2022detection, das2025efficient, khan2022anomaly} to unsupervised \cite{doshi2020continual, huang2023multi, tian2021weakly, zaheer2022generative, li2023essl}, has been explored in the literature. Beyond the training data regime, researchers have also looked at adaptive learning paradigms for VAD \cite{yao2024evaluating, doshi2022rethinking, faber2024lifelong}. The abundance of these paradigms reflects the community’s efforts to tackle VAD’s challenges from different angles. Each paradigm comes with its own assumptions, strengths, and failure modes, and part of the goal of our survey is to clarify how these pieces fit into the larger picture.

Considering the breadth of applications and methods, there is a clear need for a unifying, structured perspective on VAD. Past surveys have typically focused on a subset of this space \cite{jiao2023survey,samaila2024video, zamanzadeh2024deep, fernando2021deep, yang2023deep, ali2025anomaly, galvao2024anomaly, rezaee2024survey, fahrmann2024anomaly, mishra2024skeletal, liu2024deep, lin2025survey, santhosh2020anomaly, rathee2023automated, baccari2024anomaly, bogdoll2022anomaly, bogdoll2023perception, solaas2024systematic, olugbade2022review, ahmed2018trajectory, yang2022visual, saleh2024forest}. For instance, on algorithms for a single domain (e.g. autonomous vehicles) or on specific training paradigms (e.g., unsupervised anomaly detection). However, VAD research has now grown to encompass diverse domains and a wide array of deep learning techniques. Researchers in one domain may not be fully aware of relevant techniques developed in another domain, even though the underlying problems share similarities. We aim to bridge this gap by providing a comprehensive survey that treats VAD holistically. In particular, we bring together human-centric, vehicle-centric, and environment-centric VAD under one umbrella (see \Cref{fig:flow}). By comparing and contrasting the problem formulations, data characteristics, and successful techniques across these domains, our survey highlights common principles as well as domain-specific nuances. This unified viewpoint is intended to help transfer knowledge across application domains. Moreover, we organize the growing literature on deep learning for VAD into a coherent taxonomy, which makes it easier to understand how different approaches relate to each other. Rather than seeing the field as a collection of disjoint research focuses, readers will gain a structured map of VAD research: the key problem settings, the algorithmic families, and the connections between them. We aim to help the research community identify open problems and practical barriers that must be addressed to advance VAD towards widespread deployment. This survey aims to bring clarity to what has been accomplished and what remains to be done. We summarize not only the State-of-The-Art (SOTA) techniques but also their limitations, and we pinpoint areas ready for new exploration. To this end, the main contributions of this paper are as follow:

\begin{itemize}

    \item We identify and critically analyze key challenges and open problems in VAD. By highlighting these gaps, the survey outlines practical considerations necessary for building reliable, adaptive, and deployable VAD systems.

    \item We present a structured taxonomy of VAD approaches categorized by supervision levels and learning paradigms, including supervised, unsupervised, weakly supervised, self-supervised, and adaptive learning. This taxonomy clarifies the underlying assumptions, strengths, and limitations of each paradigm and guides readers in selecting appropriate methods for different problem settings.
    
    \item We provide a comprehensive and unified survey of VAD via deep learning, encompassing human-centric, vehicle-centric, and environment-centric domains. This work bridges the gap between fragmented subfields by systematically comparing problem formulations, data characteristics, and methods across application areas, enabling knowledge transfer and cross-domain insights.

\end{itemize}
\section{VAD Challenges}
\label{sec:challenges}

\begin{table*}[]
\centering
\caption{Comprehensive summary of key challenges in video anomaly detection. The table categorizes the challenges into six broad themes. This categorization aims to guide future research and development directions in video anomaly detection scenarios.}
\label{tab:vad_challenges}
\begin{adjustbox}{width=0.90\textwidth}
\begin{tabular}{>{\centering\arraybackslash}m{2.5cm}||>{\centering\arraybackslash}m{4cm}|>{\centering\arraybackslash}m{6.5cm}|>{\centering\arraybackslash}m{3.5cm}}
\hline
\textbf{Category} & \textbf{Challenge} & \textbf{Short Description} & \textbf{References} \\
\hline \hline

\multirow{3}{=}{\textbf{Data Scarcity and Annotation Challenges}} 
& C1: Rarity of Anomalies and Class Imbalance 
& Anomalies are inherently rare, leading to extreme class imbalance. 
& \cite{samaila2024video, ramachandra2020survey, yang2024follow}
 \\
\cline{2-4}

& C2: Limited and Difficult Labeling 
& Precise annotation is labor-intensive and often requires domain experts. 
& \cite{al2024coarse, zhu2022towards, georgescu2021anomaly}
 \\
\cline{2-4}

& C3: Ambiguity in Defining Anomalies and Context Specificity 
& Anomalies are context-dependent; similar actions can be normal or abnormal depending on scenario. 
& \cite{yang2024context, georgescu2021anomaly, narwade2025synthetic}
 \\
\hline

\multirow{3}{=}{\textbf{Spatiotemporal Modeling Challenges}} 
& C4: Complex Temporal Patterns and Long-Term Dependencies 
& Many anomalies manifest over time. Instantaneous and prolonged anomalies must be detected with appropriate temporal context. 
& \cite{tian2021weakly, wu2024vadclip, wang2021robust}
 \\
\cline{2-4}

& C5: Multi-Agent Interactions and Crowded Scenes 
& Interactions are complex, often involving crowd behavior or occlusion.
& \cite{rezaee2024survey, ponraj2025video, luo2024detecting}
 \\
\cline{2-4}

& C6: Feature Abstraction Level
& Pixel-based models are affected by visual noise. Higher abstraction may lose contextual cues. 
& \cite{alinezhad2023understanding, hirschorn2023normalizing, mishra2024skeletal, yu2023regularity, yao2024evaluating}
 \\
\hline

\multirow{3}{=}{\textbf{Robustness and Generalization Challenges}} 
& C7: Environmental Variations and Noise 
& Real-world conditions (e.g., weather, lighting) degrade model performance. 
& \cite{yang2024context, biradar2024robust, bhardwaj2025leveraging}
 \\
\cline{2-4}

& C8: Domain Shift and Cross-Scene Generalization 
& Models often fail when deployed in new visual environments.
& \cite{guo2024ada, wang2024domain, cho2024towards}
 \\
\cline{2-4}

& C9: Open-Set Nature of Anomalies and Novelty 
& Not all types of normal/anomaly can be seen during training or validation. 
& \cite{zhu2022towards, alinezhad2023understanding, noghre2024exploratory}
 \\
\cline{2-4}

& C10: Handling Concept Drift and Evolving Normality 
& Normal behavior may evolve over time; failure to adapt causes false alarms, while over-adaptation risks misclassifying anomalies. 
& \cite{georgescu2021anomaly, doshi2022rethinking, faber2024lifelong}
 \\
\hline

\multirow{3}{=}{\textbf{Evaluation and Benchmarking Challenges}} 
& C11: Scarcity of Comprehensive Benchmark Datasets 
& Benchmark datasets are limited in diversity and detail.
& \cite{liu2018ano_pred, Rodrigues_2020_WACV, danesh2023chad}
 \\
\cline{2-4}

& C12: Limitations of Current Evaluation Metrics 
& Common metrics often fail to reflect deployment performance.
& \cite{danesh2023chad, samaila2024video, alinezhad2023understanding, noghre2024exploratory}
 \\
\cline{2-4}

& C13: Gap Between Offline Evaluation and Deployment Performance 
& Real-time scenarios require new protocols for accurate assessment.
& \cite{pazho2023survey, alinezhad2023understanding, noghre2024exploratory} \\
\hline

\multirow{3}{=}{\textbf{Real-Time and Deployment Challenges}} 
& C14: Real-Time Processing and Low Latency 
& Timely detection is essential in safety-critical domains.
& \cite{nawaratne2019spatiotemporal, ali2023real, karim2024real}
 \\
\cline{2-4}

& C15: Resource Constraints and Scalability 
& Deep models require significant computational resources.
& \cite{samaila2024video, doshi2022rethinking, zhu2021video}
 \\
\cline{2-4}

& C16: Calibration and Thresholding (False Alarms vs. Misses) 
& Setting the right anomaly threshold is critical to balance false positives and false negatives.
& \cite{doshi2021online, micorek2024mulde, nie2024interleaving}
 \\
\hline

\multirow{3}{=}{\textbf{Adaptive Learning Challenges}} 
& C17: Catastrophic Forgetting or Stability-Plasticity Dilemma
& Models may lose previously learned information when updated with new data.
& \cite{ntelopoulos2024callm, asal2024ensemble, faber2024lifelong}
 \\
\cline{2-4}

& C18: Efficient Label Utilization During Adaptation 
& Labels are scarce in streaming settings.
& \cite{yao2024evaluating, doshi2022rethinking, faber2024lifelong} \\
\hline

\end{tabular}
\end{adjustbox}
\end{table*}
VAD presents unique challenges, summarized in \Cref{tab:vad_challenges}, which are explained in detail in this section.
\subsection{Data Scarcity and Annotation Challenges}

\subsubsection{\textbf{C1: Rarity of Anomalies and Class Imbalance}}

By definition, anomalies are rare events compared to normal. For instance, traffic accidents in autonomous driving are infrequent compared to normal driving scenarios. Deep learning models typically thrive on abundant data, but the scarcity of anomalous examples means they struggle to learn generalizable patterns.

\subsubsection{\textbf{C2: Limited and Difficult Labeling}}

Not only are anomalies rare, but they are also inherently difficult to label. Annotating frame-by-frame or pixel-level ground truth is labor-intensive. In many instances, expert knowledge is also essential to perform correct labeling. For example, in healthcare applications like monitoring Parkinson’s disease, identifying the exact onset and offset of anomalous behavior necessitates the involvement of domain specialists.

\subsubsection{\textbf{C3: Ambiguity in Defining Anomalies and Context Specificity}}

Unlike standard vision tasks, anomaly detection is highly context-dependent. The same behavior may be normal in one setting but anomalous in another. For instance, in public safety, punching signals violence unless occurring in a boxing gym. Such contextual ambiguity complicates defining anomalies. Modeling such context is difficult and requires auxiliary inputs or learning multiple modes of normality. While deep models must be robust to these ambiguities, current methods often struggle with subtle or context-sensitive anomalies.

 \subsection{Spatiotemporal Modeling Challenges}

 \subsubsection{\textbf{C4: Complex Temporal Patterns and Long-Term Dependencies}}

Anomalies unfold as irregular motion patterns or unusual events over time. Capturing temporal dynamics is a core difficulty. For instance, in Autonomous Driving, an accident might be inferred from a vehicle’s erratic trajectory over several seconds. Some anomalies have a slow temporal build-up (e.g., a person slowly loitering in a restricted area). Detecting these requires integrating information over long durations. On the other hand, anomalies can be instantaneous (a sudden explosion). Balancing responsiveness to quick events with the ability to analyze extended sequences is non-trivial.

\subsubsection{\textbf{C5: Multi-Agent Interactions and Crowded Scenes}}

Many anomalies involve multiple entities interacting with each other. Detecting these anomalies requires modeling collective behavior patterns. However, modeling them is difficult due to occlusions and complex dynamics. In some events the anomaly is evident in the group’s joint configuration (e.g. a group of people suddenly running away) even if each individual’s motion by itself might appear normal.

\subsubsection{\textbf{C6: Feature Abstraction Level}}

Deep video anomaly detectors traditionally operate on raw pixel data, but this raises feature redundancy issues. Raw pixel-based models must contend with background clutter, illumination changes, and camera motion that can obscure the relevant pattern. An emerging approach is to use other modalities such as pose, optical flows, object landmarks, etc. However, these approaches rely on accurate preprocessing steps. Additionally, detecting certain anomalies requires detailed visual queues that may be lost in higher levels of abstraction (e.g., detecting someone carrying a weapon would be more challenging).

\subsection{Robustness and Generalization Challenges}

\subsubsection{\textbf{C7: Environmental Variations and Noise}}

VAD methods must operate in diverse real-world conditions that can affect their inputs. Models may face day/night cycles, various weather conditions, and lighting changes. These factors can introduce visual noise that are unrelated to anomalies but can confuse deep models. Another aspect is highly dynamic backgrounds that could lead to high false alarm rates when the model interprets normal background changes as abnormal. Robustness to these perturbations is crucial.

\subsubsection{\textbf{C8: Domain Shift and Cross-Scene Generalization}}

Related to C7 is the domain shift problem: an anomaly detection model trained in one setting often fails when deployed in a new setting. This is because deep models internalize the statistics of their training data’s environment. Domain adaptation and generalization techniques are actively researched. This is a critical issue for scalability as well: a city-wide deployment across hundreds of street cameras would require per-camera calibration if the model cannot generalize.

\subsubsection{\textbf{C9: Open-Set Nature of Anomalies and Novelty}}

VAD is an open-set problem: a model can never see examples of all possible anomalies in training, since by definition anomalies encompass anything that deviates from normal, including novel events that have never occurred before. On the same note, capturing all normal behaviors is also not feasible. VAD must be prepared for the unforeseen. This translates to the “unknown unknowns” problem: an AI may handle known rare events but fail to recognize a truly odd hazard as an anomaly. The open-set challenge also complicates evaluation. A model could correctly detect all anomalies in a test set and still be unreliable in practice if a new kind of anomaly occurs.

\subsubsection{\textbf{C10: Handling Concept Drift and Evolving Normality}}

Over time, what is considered normal or even anomalous may evolve. This phenomenon is known as concept drift. In a traffic monitoring scenario seasonal differences may cause normal behavior patterns to shift. In healthcare, a patient’s baseline behavior might gradually change due to therapy or disease progression. If a model is not updated, it may raise false alarms on these evolving behaviors.

\subsection{Evaluation and Benchmarking Challenges}

\subsubsection{\textbf{C11: Scarcity of Comprehensive Benchmark Datasets}}

Most current VAD models are trained and tested on a limited set of benchmarks. While useful for initial development, these datasets often lack diversity in scenes, environmental conditions, and anomaly categories.

\subsubsection{\textbf{C12: Limitations of Current Evaluation Metrics}}

The dominant metrics used in VAD fail to fully capture the real-world effectiveness of a model. These metrics often abstract away threshold selection and ignore the impact of false alarms. Additionally, metrics rarely account for operational concerns such as alert fatigue, latency, or the cost of misclassification.

\subsubsection{\textbf{C13: Gap Between Offline Evaluation and Deployment Performance}}

Many VAD methods are evaluated in offline settings using pre-recorded video clips. Offline evaluation may overstate model accuracy. Bridging the gap between offline benchmarks and online performance requires new evaluation protocols that account for temporal causality, resource constraints, and continuous learning needs.

\subsection{Real-Time and Deployment Challenges}

\subsubsection{\textbf{C14: Real-Time Processing and Low Latency}}

For many applications, detecting anomalies promptly is crucial. For example, autonomous vehicles must detect and react to road anomalies within a very short time to avoid accidents. Such scenarios demand that deep learning models operate in real-time on video streams. Even if accuracy is high, a method that triggers an alert too late is often unacceptable in practice. Achieving real-time anomaly detection without sacrificing detection quality is an active challenge.

\subsubsection{\textbf{C15: Resource Constraints and Scalability}}

Deep learning models for video require significant memory and computation. In a real-world deployment like city-wide surveillance, running a deep anomaly detector on all feeds simultaneously is a massive scalability challenge. Likewise, an autonomous vehicle has a power and hardware budget. Thus, anomaly detection methods must be efficient in terms of computation, memory, and energy. Another aspect of scalability is handling long durations and continuous monitoring: a model might need to run 24/7. Storing and analyzing such long video sequences can be difficult. There is also a data management challenge: if anomalies are flagged often, how to store or review these events efficiently. Ensuring that a solution scales from a small benchmark to a deployment is a non-trivial jump.

\subsubsection{\textbf{C16: Calibration and Thresholding (False Alarms vs. Misses)}}

Deploying an anomaly detector in the real world requires choosing how sensitive it should be. In other words, setting thresholds or decision criteria for what is flagged as anomalous. This leads to a classic precision-recall trade-off: a very sensitive system will catch nearly all true anomalies (high recall) but at the cost of many false alarms (low precision), whereas a strict system will raise fewer false alerts but might miss anomalies. Finding the right balance is extremely challenging and often application-specific.

\subsection{Adaptive Learning Challenges}

\subsubsection{\textbf{C17: Catastrophic Forgetting or Stability-Plasticity Dilemma}}

Catastrophic forgetting is the tendency of models to overwrite previously learned knowledge when updated with new data. If a model is updated incrementally to learn from new scenes or behaviors, it may degrade in performance on previously seen data. This is critical in safety or surveillance settings, where remembering rare but significant events is essential. This challenge is closely related to the Stability-Plasticity Dilemma, which describes the trade-off between retaining existing knowledge (stability) and acquiring new knowledge (plasticity) without interference.

\subsubsection{\textbf{C18: Efficient Label Utilization During Adaptation}}

Obtaining labels in a streaming setting is expensive and time-consuming. Therefore, continual learning must proceed with minimal supervision. Designing models that can effectively leverage sparse and noisy labels, or self-supervise their adaptation process, is a key challenge.
\section{Definitions and Solutions}
\label{sec:paradigms}
\begin{figure}[]
\centering
\includegraphics[clip,trim={55 55 55 75},width=0.8\linewidth]{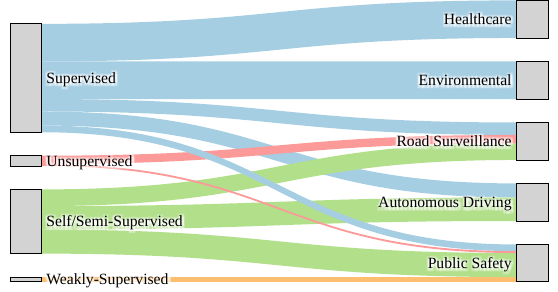}
\caption{Percentage distribution of supervision levels within each domain.}
\label{fig:sankey}
\end{figure}

VAD focuses on identifying patterns or events in video sequences that deviate significantly from expected or normal behavior \cite{pazho2023survey}. As discussed in \Cref{sec:intro}, the complexity of VAD has led researchers to adopt varying levels of supervision (see \Cref{fig:sankey}). This section identifies and discusses the definition and solutions within each supervision level. \Cref{tab:solutions} summarizes the solutions and their weaknesses and strengths. 

\subsection{Supervised VAD}

Supervised VAD involves training models on labeled datasets where both normal and anomalous events are explicitly annotated. This approach is particularly effective in domains where anomalies are well-defined and annotated data is available, such as in healthcare. By learning from labeled examples, supervised methods can achieve high accuracy in detecting known types of anomalies. However, as outlined in \Cref{sec:challenges} due to challenges such as Rarity of Anomalies and Class Imbalance, Limited and Difficult Labeling, Ambiguity in Defining Anomalies and Context Specificity, and Open-set nature of Anomalies and Novelty (challenges C1, C2, C3, and C9) supervised approaches exhibit limited applicability \cite{pazho2023survey, cai2025medianomaly, darban2024dacad, samaila2024video, ramachandra2020survey}. The main solution in this supervision level is treating VAD as a classification problem.

\textbf{Supervised Classification (S1):} The most common formulation of supervised VAD is as a classification task, where models are trained to distinguish between predefined normal and anomalous events. This setup leverages well-established classification algorithms to learn discriminative features.

\subsection{Weakly-Supervised VAD}

To address the challenge of obtaining accurately labeled data for supervised solutions (challenges C2 and C3), weakly-supervised approaches offer greater flexibility. In these methods, labels may be incomplete, noisy, or ambiguous. Weakly-supervised solutions mostly take advantage of Multiple Instance Learning (MIL) and try to improve it for better efficacy.

\begin{figure}[t]
\centering
\includegraphics[clip,trim={18 18 18 18},width=0.85\linewidth]{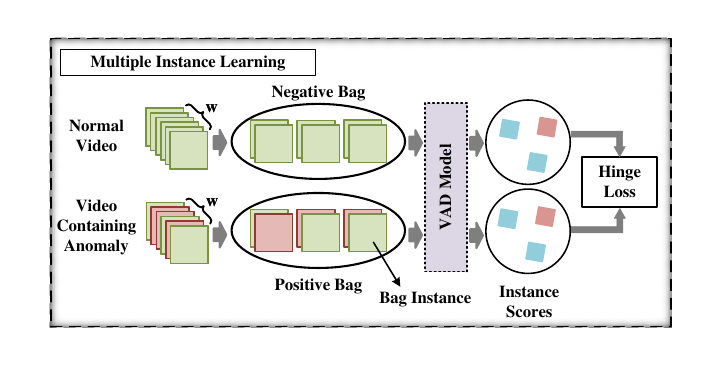}
\caption{VAD formulated as a weakly supervised problem, commonly addressed using MIL (S2).}
\label{fig:MIL}
\end{figure}

\textbf{Multiple Instance Learning (S2)}: MIL treats each video as a bag of instances, labeling it anomalous if at least one instance is abnormal (see \Cref{fig:MIL}). During training, the model learns to identify which instances within positive bags are anomalous, without needing fine-grained labels. One-stage MIL often focuses on the most prominent anomaly, risking missed detections of subtle instances, while two-stage self-training methods use MIL to generate pseudo-labels iteratively, refining both the model and labels, enabling more robust and comprehensive detection of both obvious and subtle anomalies.

\subsection{Self/Semi-supervised VAD}

Semi-supervised solutions bridge supervised and unsupervised learning paradigms by using only normal videos during training to learn the characteristics of normal behavior. Previous literature often classified these methods as unsupervised. However, recent works \cite{zaheer2022generative, li2023essl} have reclassified them as semi-supervised due to the inherent supervision involved: normal and abnormal sequences are distinguished, and only normal sequences are utilized during training. This shift in terminology acknowledges the partial labeling and guidance provided, which differentiates these methods from truly unsupervised approaches. In general, most of these methods also fall under the self-supervised paradigm, where supervisory signals are derived from inherent characteristics of the normal data. Depending on the learning objective, these solutions can be categorized into four main groups: Reconstruction-based, Prediction-based, Jigsaw Puzzle, and Distribution Estimation.

\begin{figure}[t]
\centering
\includegraphics[clip,trim={18 25 18 25},width=1\linewidth]{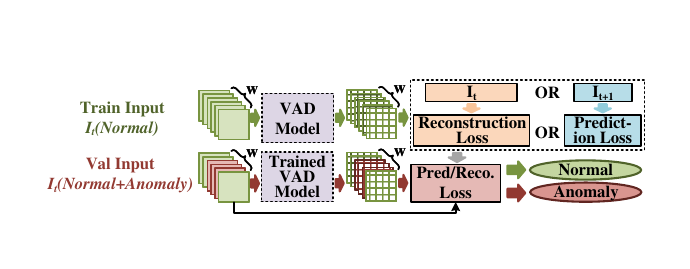}
\caption{Self/semi-supervised VAD achieved through reconstruction(S3) or prediction (S4). The top figure illustrates the training phase using only normal data, while the bottom shows the inference phase, where elevated loss indicates abnormal behavior. }
\label{fig:rec_pred}
\end{figure}

\textbf{Reconstruction-based (S3):} This strategy employs autoencoders to reconstruct normal data; anomalies are indicated by high reconstruction loss when the model fails to reconstruct anomalous snippets accurately, as seen in \Cref{fig:rec_pred}.

\begin{figure*}[t]
\centering
\includegraphics[clip,trim={18 18 18 18},width=0.85\textwidth]{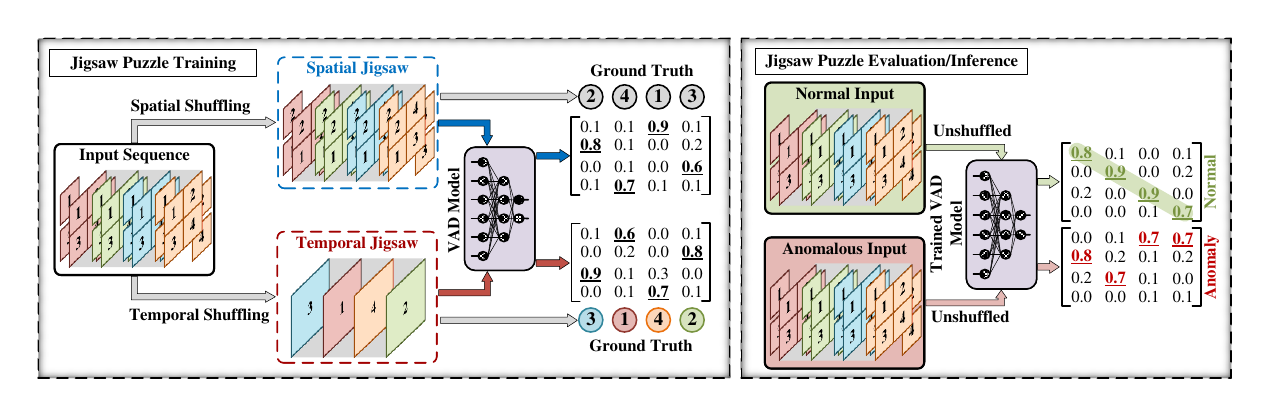}
\caption{Self/semi-supervised VAD achieved through jigsaw puzzle task (S5). The puzzle can be spatial, temporal, or a combination. The left figure illustrates the training phase using only normal data, while the right shows the inference phase, where wrong permutation prediction indicates abnormal behavior.}
\label{fig:jig}
\end{figure*}
\textbf{Prediction-based (S4):} In these approaches, models are trained to predict the future normal behavior, with anomalies identified through higher prediction loss on abnormal sequences, as seen in \Cref{fig:rec_pred}.

\textbf{Jigsaw Puzzle (S5):} A supervisory signal is generated by formulating a jigsaw puzzle task, which may be spatial, temporal, or a combination of both (see \Cref{fig:jig}). The model is trained exclusively on normal data, learning to reassemble shuffled video segments. During inference, its ability to correctly reconstruct these sequences is used as a measure for computing anomaly scores.

\textbf{Distribution Estimation (S6):} This category employs either non-deep learning or deep learning methods to model the distribution of normal samples during training. At inference, instances with low likelihood are identified as anomalies.
\begin{figure}[]
\centering
\includegraphics[clip,trim={18 18 18 18},width=0.85\linewidth]{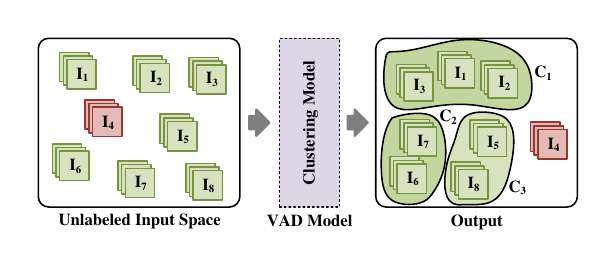}
\caption{Unsupervised anomaly detection through clustering.  }
\label{fig:clustering}
\end{figure}
\subsection{Unsupervised} 
In unsupervised training, no labels are available to distinguish between normal and anomalous instances. However, the literature frequently misclassifies certain self-supervised or semi-supervised approaches as unsupervised. A critical observation is that many of these methods are trained exclusively on normal data. This implicitly introduces label information, violating the core principle of unsupervised learning \cite{zaheer2022generative}. Consequently, such models should not be considered unsupervised. The degree of supervision must be evaluated not only based on the methodology but also in relation to the informational content embedded in the training data. Despite its fundamental nature, fully unsupervised anomaly detection remains relatively underexplored compared to its self-supervised and semi-supervised counterparts, indicating a significant opportunity for advancement and application in real-world scenarios.

Truly unsupervised methods operate without any access to ground-truth labels. These approaches aim to exploit the normality advantage; the observation that anomalies represent rare and irregular events, whereas the majority of the data corresponds to normal behavior \cite{wang2019effective, yu2022deep}. The core strategy behind these methods is to leverage this statistical imbalance: given that normal samples dominate the dataset, the global structure and distributional trends of the data are expected to reflect normal characteristics. Unsupervised models are therefore trained to capture these prevailing patterns, under the assumption that deviations from the learned representation will correspond to anomalous instances.

\begin{table*}[b]
\centering
\caption{Overall classification of VAD solutions.}
\label{tab:solutions}
\begin{adjustbox}{width=0.85\textwidth}
\begin{tabular}{@{}>{\centering\arraybackslash}m{2cm}|
                >{\centering\arraybackslash}m{2.5cm}
                >{\centering\arraybackslash}m{4cm}
                >{\centering\arraybackslash}m{4cm}
                >{\centering\arraybackslash}m{3.5cm}@{}}
\toprule
\textbf{Supervision Level} & \textbf{Solution}              & \textbf{Definition}                                                                    & \textbf{Main Strength}                                                                          & \textbf{Main Limitation}                                                                        \\ 
\midrule \midrule
\textbf{Supervised}        & S1: Supervised Classification  & Frames VAD as a classification task using labeled datasets.                            & High accuracy and reliability in detecting predefined, labeled anomalies.                       & Requires extensive and exhaustive labeled datasets, often impractical for real-world scenarios. \\ 
\midrule \midrule 
\multirow{2}{*}{\parbox{2.5cm}{\centering \textbf{Weakly-}\\\textbf{Supervised}}} 
                           & S2: Multiple Instance Learning & Labels video bags; identifies anomalous instances.                                     & Can handle weak labels where only bag-level annotation is provided, reducing labeling efforts.  & Often overlooks subtle or less prominent anomalies, leading to higher false-negative rates.     \\ 
\midrule \midrule
\multirow{5}{*}{\parbox{2.5cm}{\centering \textbf{Self/}\\\textbf{Semi-}\\\textbf{Supervised}}}   
                           & S3: Reconstruction             & Uses autoencoders to reconstruct normal data; anomalies have high reconstruction loss. & Effective for capturing the structure of normal behaviors.                                      & Struggles with generalization when normal patterns exhibit high variability.                    \\ \cmidrule(l){2-5} 
                           & S4: Prediction                 & Predicts future behavior; high prediction loss indicates anomalies.                    & Performs well on temporal datasets by learning sequential patterns.                             & Suffers in scenarios with non or limited temporal context availability.                         \\ \cmidrule(l){2-5} 
                           & S5: Jigsaw Puzzle              & Challenges models to reassemble shuffled video segments.                               & Improved detection of subtle anomalies.                                                         & Complexity of solving permutations in jigsaw puzzles, affecting real-time VAD.                  \\ \cmidrule(l){2-5} 
                           & S6: Distribution Estimation    & Uses generative models or statistical methods to learn normal behavior distributions.  & Captures distributional properties of normal behaviors.                                         & Sensitive to noise.                                                                             \\
\midrule \midrule
\multirow{2}{*}{\textbf{Unsupervised}}     
                           & S7: Clustering                 & Uses clustering techniques.                                                            & No reliance on labeled data and simple implementation.                                          & Limited generalizability, sensitive to hyperparameters.                                         \\ \cmidrule(l){2-5} 
                           & S8: Pseudo-Label Induction     & Leverage error magnitude to do pseudo-labeling for filtering anomalies.                & No reliance on labeled data.                                                                    & Pseudo labels are uncertain and can potentially reinforce false patterns.                       \\
\bottomrule
\end{tabular}
\end{adjustbox}
\end{table*}

\textbf{Clustering (S7):} Clustering methods assume that normal data form dense clusters in feature space, while outliers in low-density regions are potential anomalies, as illustrated in \Cref{fig:clustering}. Approaches range from classical algorithms like k-means to deep clustering methods that jointly learn features and clusters. Despite their effectiveness, clustering methods face challenges such as sensitivity to hyperparameters, such as the number of clusters, and reliance on clear structural differences between normal and anomalous data.

\textbf{Pseudo-label Induction (S8)}: This strategy leverages the normality assumption: normal data dominate the input distribution. While conceptually related to self or semi-supervised approaches, a key distinction is that the training data includes unknown anomalies. These methods use reconstruction errors or prediction inconsistencies to assign pseudo-labels, guiding anomaly filtering or classifier training. As they rely on self-generated signals without ground truth, they are considered unsupervised self-supervised approaches. However, unreliable pseudo-labels and feedback loops can undermine robustness and generalizability, especially in noisy or complex data.
\section{Adaptive Learning in VAD}

As discussed in \Cref{sec:intro} and \Cref{sec:challenges}, VAD is a dynamic and complex problem, ever evolving and heavily affected by spatio-temporal changes. This includes but is not limited to Environmental Variations and Noise, Domain Shift, Open-Set Nature, Concept Drift, and Calibration and Thresholding (challenges C7, C8, C9, C10, and C16). To address these challenges, adaptive learning methods such as meta-learning, online learning, continual learning, and active learning have become essential \cite{lu2018learning, saurav2018online, wu2020adversarial, tang2020deep}. In this survey, the term "adaptive learning" encompasses a range of general adaptation methods. These techniques enable models to update and adjust to new data, trying to manage the aforementioned challenges.

\textbf{Meta-learning}, also known as "learning to learn," \cite{thrun1998learning} focuses on designing models capable of rapidly adapting to new tasks by leveraging knowledge acquired from previous tasks. This approach involves training across a variety of tasks to develop a general learning strategy, enabling the model to perform effectively on novel tasks with minimal data, which is particularly useful for solving the domain shift problems in VAD models. One significant weakness of meta-learning, particularly in real-time VAD, is the high computational expense associated with training across multiple tasks. However, integrating meta-learning with few-shot training methods can help mitigate this issue by enabling the model to learn from a limited number of examples, thereby reducing the computational burden while maintaining adaptability and performance. \cite{lu2020few} introduces a meta-learning framework using the Model-Agnostic Meta-Learning (MAML) \cite{finn2017model} algorithm to enhance semi/self-supervised anomaly detection in surveillance videos. This approach involves training the model on various scenes, creating tasks that simulate few-shot learning scenarios. 

\textbf{Online Learning} is a paradigm where the model is updated incrementally as it receives new data points \cite{hazan2016introduction, hoi2021online}. This approach allows the model to adapt continuously to new information. Online learning is particularly advantageous when dealing with large datasets or streaming data, as it can handle data efficiently without requiring access to the entire dataset simultaneously. Online learning has been explored for anomaly detection on other types of data, such as time series \cite{saurav2018online}, text \cite{9330769}, and medical images \cite{chen2023effective}. In VAD, Yao et al. \cite{Yao_2024_CVPR} introduced a framework optimized for real-world deployment, integrating inference and training in a pipeline to enhance public safety applications. While effective under conditions of minimal distributional shift, online learning faces notable limitations. These include susceptibility to noisy or unrepresentative data (challenges C1 and C7), as well as challenges such as the stability-plasticity dilemma and catastrophic forgetting (challenge C17), where frequent updates may overwrite prior knowledge.

\textbf{Continual Learning} is a strategy in machine learning where a model is designed to continually acquire, fine-tune, and retain knowledge from a stream of data over an extended period. This approach addresses the challenge of catastrophic forgetting (challenge C17), where learning new information can lead to a loss of previously acquired knowledge \cite{french1999catastrophic, mccloskey1989catastrophic}. This enables models to adapt to new tasks and changes in data distribution without sacrificing performance on previously learned tasks, making it particularly valuable in dynamic environments where the data evolves over time. Continual learning encounters challenges related to managing the high volume of streaming data and maintaining the efficiency of continuous model training. That is why most of the works in this area move toward few-shot learning to be able to handle the complexity of the training process while making real-time decisions. \cite{doshi2022rethinking} proposed a two-step method for anomaly detection using deep learning-based feature extractors combined with kNN and a memory module, enhanced by two continual learning approaches. The first approach involves exact k-Nearest Neighbor kNN distance computation, effective for incrementally learning nominal behaviors when the training data size is manageable, updating the memory module with kNN distances from each training split. To address the computational expense as the training set grows, the second approach employs a fully connected deep neural network (kDNN) to estimate kNN distances, ensuring scalability and efficiency.

\textbf{Active Learning} \cite{cohn1994improving} is a technique where the algorithm selectively queries the user (or domain experts) to label new data points to improve the learning efficiency and model performance \cite{monarch2021human, settles2009active}. In scenarios where labeling data is costly or time-consuming, active learning is particularly valuable because it allows the model to focus on acquiring labels for the most informative data points. This is achieved through various strategies that prioritize data points based on criteria such as uncertainty, representativeness, or expected model change \cite{settles2009active}. By enabling the model to query the most useful data points for annotation, active learning reduces the need for large pre-labeled data and enhances the model's ability to generalize from fewer labeled instances (challenges C1, C2, and C18). Incorporating human feedback on selected samples within an active learning framework establishes a few-shot learning paradigm that improves the efficacy of anomaly detection systems and the efficiency of training the model. A significant challenge associated with this technique is the requisite involvement of a human or domain expert (challenges C2 and C18). This requirement can introduce complexities related to scalability and efficiency, as the continuous need for expert input can limit the speed and autonomy of the learning process. \cite{doshi2020continual} proposes an active learning framework using YOLO v3 \cite{redmon2016you} and Flownet 2 \cite{ilg2017flownet} for feature extraction and kNN for anomaly detection. The model constructs a statistical baseline of normal behaviors using kNN distances and continually updates it with new nominal data. Anomalies trigger human feedback for labeling, which categorizes this work as an active learning framework rather than continual learning, as described in the original paper. Several other works propose a more advanced method for selecting the data queries. \cite{loy2010stream} proposes an adaptive weighting scheme for dynamically selecting between various criteria such as the likelihood criterion, which selects samples with low likelihood according to the current model to discover new classes, and the uncertainty criterion, which selects samples that cause the most disagreement among committee members to refine the decision boundary. \cite{loy2012stream} utilizes a Bayesian nonparametric model, specifically the Pitman-Yor Process (PYP) for managing imbalanced class distributions (challenge C1) and models probabilities for both known and unknown classes.

\section{Human-Centric VAD}

\begin{table*}[]
\centering
\caption{Reviewed works in healthcare: all studies employ supervised learning; * denotes studies that evaluate multiple architectures.}
\label{tab:healthcare}
\begin{adjustbox}{width=0.90\textwidth}
\begin{tabular}{>{\centering\arraybackslash}m{2cm}|
                >{\centering\arraybackslash}m{1.25cm}|
                >{\centering\arraybackslash}m{2.5cm}|
                >{\centering\arraybackslash}m{8.5cm}|
                >{\centering\arraybackslash}m{2.5cm}}
\hline
\textbf{Task} & \textbf{Approach} & \textbf{Architecture} & \textbf{Distinct Characteristics / Novel Contributions} & \textbf{Modalities} \\
\hline\hline
\multirow{8}{*}{\parbox{2cm}{\centering Fall\\Detection}}
& \cite{chen2020vision} & CNN, LSTM & Performs person detection and contour-based feature extraction, followed by attention-guided LSTM & RGB \\ \cline{2-5}
& \cite{cai2020vision} & CNN & Mitigates feature loss through multi-task learning and leverages latent features for decision-making & RGB \\ \cline{2-5}
& \cite{chhetri2021deep} & MLP & Proposes enhanced optical dynamic flow for improved temporal motion estimation in fall scenarios & Optical Flow \\ \cline{2-5}
& \cite{chen2020fall} & Heuristic Rule-based Model & Uses pose-based features to compute fall index based on body posture changes & Pose \\ \cline{2-5}
& \cite{keskes2021vision} & GCN & Constructs a spatiotemporal graph of human poses and applies graph convolution & Pose \\ \cline{2-5}
& \cite{wang2020fall} & Random Forest, MLP & Divides falls into dynamic/static states; uses fusion of vision-based data & RGB, Pose \\ \cline{2-5}
& \cite{zhang2020human} & CNN, Logistic Regression & Models body dynamics with an inverted pendulum and analyzes motion stability to extract features & RGB, Pose \\ \cline{2-5}
& \cite{khraief2020elderly} & CNN & A multi-stream CNN where each stream processes different features & RGB, Depth, Optical Flow \\

\hline \hline
\multirow{9}{*}{\parbox{2cm}{\centering Parkinson's\\Detection}}
& \cite{gomez2021improving} & Deep Residual Network & A multimodal system using facial features and expression-specific actions for effective detection & RGB (Facial Video) \\ \cline{2-5}
& \cite{gomez2012exploring} & CNN, SVM & Analyzes evoked facial expressions using domain adaptation from face recognition & RGB (Facial Video) \\ \cline{2-5}
& \cite{gong2020novel} & CNN, SVM & Uses gait energy images to classify Parkinson's gait leveraging one-class SVM & Gait \\ \cline{2-5}
& \cite{kaur2022vision} & * & Predicts gait dysfunction by extracting 2D poses, reconstructing 3D gait from multiviews, and analyzing features using classical and deep learning models & Gait \\ \cline{2-5}
& \cite{connie2022pose} & Random Forest & Analyzes stride variability and cadence using pose-based features for effective detection & Pose \\ \cline{2-5}
& \cite{jin2020diagnosing} & * & Identifies Parkinson's symptoms via jitter and amplitude of small muscle groups in face videos & Facial Landmarks \\ \cline{2-5}
& \cite{monje2021remote} & * & Performs remote assessment using webcam video by extracting hand landmarks & Hand Landmarks \\ \cline{2-5}
& \cite{archila2022multimodal} & CNN, Random Forest & Fuses eye-tracking and gait data using covariance descriptors for Parkinson progression quantification & RGB (Eye Video), Gait \\

\hline \hline
\multirow{8}{*}{\parbox{2cm}{\centering Autism\\Detection}}
& \cite{sun2020spatial} & 3D CNN, LSTM & Proposes spatial attentional bilinear pooling to capture fine-grained spatial features and dynamic attention on discriminative regions & RGB \\ \cline{2-5}
& \cite{chen2019attention} & CNN, LSTM & Integrates photo-taking and image-viewing modalities through multi-modal knowledge distillation, enabling accurate detection using temporal and attentional behavioral features & RGB \\ \cline{2-5}
& \cite{jiang2017learning} & CNN, SVM & Analyzes attention pattern differences using discriminative image selection and fixation maps, followed by linear SVM classification & RGB (Eye Video) \\ \cline{2-5}
& \cite{tao2019sp} & CNN, LSTM & Extracts visual and temporal features from gaze scanpaths using saliency-guided patch extraction for sequence-based prediction & Scanpath \\ \cline{2-5}
& \cite{ali2022video} & 3D-CNN & Utilizes 3D-CNN for spatiotemporal analysis, focusing on action recognition to detect symptoms & RGB, Optical Flow \\ \cline{2-5}
& \cite{wu2021machine} & CNN, MLP & Processes facial expressions for autism screening & RGB, Facial Landmarks \\ \cline{2-5}
& \cite{kojovic2021using} & LSTM & Focuses on posture and movement data in social interactions for detection & Pose \\

\hline \hline
\multirow{9}{*}{\parbox{2cm}{\centering Seizure\\Detection}}
& \cite{yang2021video} & CNN, LSTM & Analyzes spatial vs. spatiotemporal features for detection, showing the latter performs better & RGB \\ \cline{2-5}
& \cite{hou2022self} & Transformer & Applies BART-inspired self-supervised training on hospital videos to learn context, followed by classification for seizure detection & RGB \\ \cline{2-5}
& \cite{pothula2022real} & * & Emotion detection used as a feature extractor & RGB (Facial Video) \\ \cline{2-5}
& \cite{ahmedt2019vision} & CNN, LSTM & Reconstructs 3D facial geometry to capture mouth and cheek motions, extracts spatial features from a defined ROI, and uses an LSTM to model temporal dynamics for seizure classification & RGB (Facial Video) \\ \cline{2-5}
& \cite{chou2023convolutional} & CNN & Transforms EEG into second-order Poincaré plots and uses pre-trained CNNs to classify seizure stages & RGB (EEG Video) \\ \cline{2-5}
& \cite{garccao2023novel} & SVM & Applies dimensionality reduction techniques (PCA and ICA), and defines handcrafted features for a SVM classifier & Optical Flow \\ \cline{2-5}
& SETR \cite{mehta2023privacy} & Transformer & Uses pretrained networks for spatial features, a transformer for temporal modeling, and Progressive Knowledge Distillation for early detection & Optical Flow \\ \cline{2-5}
& \cite{ahmedt2019motion} & CNN & Generates a compact image representation capturing the location variance and periodicity of semiology & RGB, Optical Flow \\ \cline{2-5}
& \cite{hou2021multi, hou2022automated} & GCN, TCN & A multistream framework leveraging GCN, spatio-temporal feature extraction, and late fusion & RGB, Pose, Facial Landmarks \\
\hline
\end{tabular}
\end{adjustbox}
\end{table*}

\subsection{Healthcare}

In healthcare VAD, the goal is to detect deviations in physiological or behavioral patterns that may signal disease, injury, or other medical conditions, enabling early diagnosis and intervention to improve outcomes and reduce costs. These systems might process various data types, but in this work, we focused on methods that use video as their primary data. As shown in \Cref{fig:flow}, most approaches adopt a supervised learning paradigm, reflecting the domain's need for precise, reliable detection. Research primarily targets events with strong visual cues, such as falls, Parkinson’s episodes, autistic behaviors, and seizures, which exhibit distinctive motion or posture patterns amenable to visual analysis (see \Cref{tab:healthcare}).

\subsubsection{\textbf{Fall Detection}}
Fall detection is a key task in healthcare-related video analysis due to its distinct visual patterns and practical significance. Early methods typically use RGB video and leverage pre-trained models, applying object detection and temporal modeling to track human motion. For instance, \cite{chen2020vision} uses an LSTM to distinguish fall-like behaviors over time, while \cite{cai2020vision} proposes a two-stage approach with a convolutional autoencoder for feature extraction, followed by a lightweight classifier for final prediction.

To address the limitations of RGB-only approaches, particularly under challenging conditions such as poor lighting, occlusions, or background clutter, recent works have incorporated additional modalities to improve fall detection performance. Optical flow captures pixel-level motion between frames, offering a richer representation of dynamic events; for instance, \cite{chhetri2021deep} uses optical flow with a fine-tuned VGG16 network \cite{simonyan2014very} to enhance motion-specific feature learning. Human pose estimation further improves robustness by abstracting subjects into skeletal representations, which are less sensitive to visual noise. Pose-based features such as centroid velocity and rotational energy have been applied using both deep learning and traditional classifiers, including logistic regression \cite{zhang2020human}, and hybrid models like the Multi Layer Perceptron (MLP) combined with random forest in \cite{wang2020fall}. Advancing beyond handcrafted descriptors, \cite{keskes2021vision} introduces a spatiotemporal graph convolutional network (ST-GCN) for end-to-end learning of pose dynamics. To further enhance robustness and capture complementary information, some studies combine multiple modalities such as RGB, depth maps, optical flow, and motion history images, processed through specialized network branches \cite{khraief2020elderly}.

\subsubsection{\textbf{Parkinson Detection}}

Parkinson’s disease (PD) exhibits both motor and non-motor symptoms, with motor manifestations such as tremor, rigidity, bradykinesia, postural instability, and shuffling gait being the most visually detectable and thus well-suited for computer vision analysis. Leveraging this visual accessibility, recent research has focused on facial and body movement analysis to identify Parkinsonian signs. A key facial symptom, hypomimia (reduced expressiveness) has been widely studied. For instance, \cite{jin2020diagnosing} applies facial landmark detection to extract handcrafted features classified with traditional algorithms, while \cite{gomez2021improving} enhances facial analysis through segmentation and hybrid learning strategies. More recent end-to-end approaches, such as \cite{gomez2012exploring}, repurpose pretrained face recognition models via transfer learning to detect PD and assess motor impairment severity using multiple Support Vector Machine (SVM) classifiers.

Another line of research focuses on gait and pose-based analysis, targeting motor irregularities such as bradykinesia (slowness and reduced movement amplitude) common in PD. For example, \cite{monje2021remote} uses hand keypoint trajectories during motor tasks, classifying temporal patterns with conventional models like logistic regression and random forests. Other studies analyze full-body motion through silhouettes or skeletal poses; \cite{gong2020novel} creates Gait Energy Images (GEIs), while \cite{connie2022pose} applies pose estimation followed by SVM classification. A more advanced approach by \cite{kaur2022vision} combines multi-view RGB video with 3D skeletal reconstruction and deep models, including multi-scale residual networks, achieving strong generalization. Some studies have sought to combine multiple modalities, such as facial and body movement cues, to enhance detection robustness. For instance, \cite{archila2022multimodal} proposes a multimodal framework that integrates facial expressions and skeletal motion features, aiming to capture complementary signals associated with Parkinsonian motor deficits.

\subsubsection{\textbf{Autism Detection}}
Autism Spectrum Disorder (ASD) is a common neurodevelopmental condition in children, marked by social communication deficits and atypical attention patterns. Clinical assessment relies on repeated, time-intensive behavioral evaluations by trained professionals, which are prone to subjective variability. As a result, developing automated, objective tools for ASD detection is critical to enable early, consistent, and scalable diagnosis.

One research direction leverages eye gaze patterns as behavioral biomarkers for ASD, given their link to impaired social engagement and disruptions in the social brain network. Jiang et al. \cite{jiang2017learning} used VGG-16 to analyze fixation difference maps and classified visual attention features with a linear SVM. Chen and Zhao \cite{chen2019attention} combined ResNet-50 \cite{he2016deep} with LSTM layers to model spatial-temporal gaze dynamics. Tao et al. \cite{tao2019sp} proposed SP-ASDNet, using saliency maps from neurotypical individuals to guide patch selection, followed by a CNN-LSTM network to detect deviations indicative of ASD.

Beyond gaze, many studies focus on general behavioral patterns, especially stereotypical behaviors like clapping, arm flapping, and repetitive movements, common indicators in ASD diagnosis. Ali et al. \cite{ali2022video} use 3D CNNs to detect such actions, supporting clinical assessments without providing a final diagnosis. Wu et al. \cite{wu2021machine} offer a more integrated pipeline, combining deep models on RGB and facial landmarks with statistical features (e.g., behavior frequency and duration), fed into a neural network for classification, linking low-level behavior detection with high-level diagnostic inference.

A more recent, data-driven approach eliminates manual feature engineering by end-to-end deep learning models that learn discriminative patterns directly from video. Sun et al. \cite{sun2020spatial} combine CNNs with LSTMs to extract spatial-temporal features from pixel data, while Kojovic et al. \cite{kojovic2021using} use a similar architecture with human pose inputs, offering a more abstract and potentially robust representation. 

Together, these studies form a continuum from explicit behavior modeling to implicit feature learning, highlighting the progression toward more generalizable and efficient systems. Each category of methods, whether based on gaze analysis, stereotyped motor behavior, or end-to-end learning, addresses different aspects of the complex behavioral phenotype associated with ASD, and collectively, they underscore the potential of machine learning in revolutionizing autism diagnosis.

\subsubsection{\textbf{Seizure Detection}}

Seizure detection has traditionally relied on Electroencephalography (EEG), often paired with video (VEEG) to link motor behaviors with brain activity. Some works, such as \cite{chou2023convolutional}, convert EEG data into visual forms like Poincaré plots for classification via pre-trained CNNs. While effective, EEG remains intrusive and impractical for long-term or ambulatory use. As a result, recent efforts have focused on video-only systems that analyze visible cues such as facial expressions and body movements, offering non-invasive, scalable, and more comfortable alternatives.

Building on this shift, recent work has explored methods focusing on facial features and expressions, particularly facial semiology (e.g., involuntary movements). Pothula et al. \cite{pothula2022real} use standard facial recognition pipelines to extract features for classification, while Ahmedt-Aristizabal et al. \cite{ahmedt2019vision} enhance this by modeling 3D facial dynamics, especially mouth motion, using LSTM networks. 

Another research direction focuses on full-body movement, which is more pronounced in generalized seizures. Yang et al. \cite{yang2021video} use CNNs and LSTMs to capture spatial and temporal motion features. More recent work, such as Hou et al. \cite{hou2022self}, introduces transformer-based models with BART-style self-supervised pretraining, enabling effective seizure classification with reduced dependence on large labeled datasets.

To address privacy concerns in video-based seizure monitoring, recent studies use de-identified features such as optical flow, which captures motion without revealing identity. Garção et al. \cite{garccao2023novel} apply dimensionality reduction and SVMs to optical flow, while Mehta et al. \cite{mehta2023privacy} propose a CNN-transformer hybrid with Progressive Knowledge Distillation for early prediction. Complementing this, multimodal fusion strategies have been explored to improve detection robustness. Ahmedt-Aristizabal et al. \cite{ahmedt2019motion} integrate facial and hand movements to create compact semiology descriptors, while Hou et al. \cite{hou2021multi, hou2022automated} fuse RGB, optical flow, body pose, and facial landmarks via multi-branch networks to produce richer representations for seizure classification.

\begin{table*}[]
\centering
\caption{Overview of Pixel-based approaches in VAD for public safety. * denotes that multiple alternative architectures have been used.}
\label{tab:pixel_based_table}
\begin{adjustbox}{width=0.9\textwidth}
\begin{tabular}{>{\centering\arraybackslash}m{2cm}|
                >{\centering\arraybackslash}m{2.5cm}|
                >{\centering\arraybackslash}m{1cm}|
                >{\centering\arraybackslash}m{2cm}|
                >{\centering\arraybackslash}m{7cm}|
                >{\centering\arraybackslash}m{1.5cm}}
\hline
\textbf{Approach} & \textbf{Supervision} & \textbf{Strategy} & \textbf{Architecture} & \textbf{Distinct Characteristics / Novel Contributions} & \textbf{Modality} \\
\hline \hline

\cite{kirichenko2022detection} & Supervised & S1 & CNN, RNN & Using CNN as spatial feature extractor and RNN for temporal pattern detection and final classification & Pixel \\ \hline
\cite{martinez2021criminal} & Supervised & S1 & 3D CNN & Use 3D CNN for simultaneous spatiotemporal analysis & Pixel \\ \hline
\cite{muneer2023shoplifting} & Supervised & S1 & CNN, LSTM & Uses Inception V3 blocks and LSTM for feature extraction & Pixel \\ \hline
ADOS \cite{manikandan2022neural} & Supervised & S1 & CNN & Minimizes multi-object detection errors by segmenting frames and applying a saliency-aware classification  & Pixel \\ \hline
\cite{das2025efficient} & Supervised & S1 & * & Two-stage gun detection using fine-tuned spatial classifiers and temporal sequence models & Pixel \\ \hline
\cite{bhatti2021weapon} & Supervised & S1 & CNN & Uses off-the-shelf object detectors and reduces false positives by incorporating confusion classes & Pixel \\ \hline
\cite{nyajowi2021cnn} & Supervised & S1 & CNN, LSTM & CNN–LSTM deep learning model that combines spatial feature extraction from convolutional layers with temporal sequence modeling from LSTM & Pixel \\ \hline \hline

\cite{yang2019enhanced} & Self/Semi-supervised & S4 & CNN & Dual-branch model with adversarially enhanced reconstruction and object-focused scoring based on likelihood, position, and confidence & Pixel \\ \hline
NM-GAN \cite{chen2021nm} & Self/Semi-supervised & S4 & CNN & Uses noise-modulated adversarial learning, where a discriminator trained on noise-injected reconstruction errors distinguishes normal from anomalous patterns & Pixel \\ \hline
\cite{georgescu2021anomaly} & Self/Semi-supervised & S5, S6 & CNN, 3D CNN & Defining multiple tasks—arrow of time prediction, motion shuffling, irregularity prediction (viewed as various jigsaw puzzles), and knowledge distillation & Pixel \\ \hline
\cite{barbalau2023ssmtl++} & Self/Semi-supervised & S5, S6 & CNN, 3D CNN & Adds adversarial pseudo anomalies, segmentation, jigsaw, pose estimation, and inpainting to multi-task training & Pixel \\ \hline
\cite{wang2022video} & Self/Semi-supervised & S6 & 3D CNN & Decoupled spatial and temporal jigsaw puzzles and employed a multi-label paradigm for more accurate VAD & Pixel \\ \hline
\cite{georgescu2021background} & Self/Semi-supervised & S4 & CNN & Uses pseudo-abnormal examples to guide the autoencoder and binary classifiers for each branch & Pixel, Flow \\ \hline
\cite{huang2023multi} & Self/Semi-supervised & S5 & CNN & A dual-encoder single-decoder model that aligns appearance and flow features and uses memory of normal prototypes to enhance detection accuracy & Pixel, Flow \\ \hline
\cite{luo2021future} & Self/Semi-supervised & S5 & CNN &Future prediction is guided by flow, intensity, and gradient losses, with a discriminator improving frame realism\\ \hline
\cite{doshi2020continual} & Semi-Supervised & S7 & - & Combines flow and object detections to form feature vectors for statistical anomaly detection & Pixel, Flow \\ \hline \hline

ARMS \cite{shi2023abnormal} & Weakly-supervised & S2 & CNN & Trained through bootstrapped pseudo labeling, hard anomaly mining, and adaptive self-training with dynamic abnormal ratios to capture both easy and subtle anomalies & Pixel \\ \hline
RTFM \cite{tian2021weakly} & Weakly-supervised & S2 & CNN & Assumes abnormal snippets have higher feature magnitudes, selects top-k segments per video to maximize abnormal-normal separation & Pixel \\ \hline
\cite{li2022attention} & Weakly-supervised & S2 & Transformer & Uses ViLBERT to extract multimodal features per segment, followed by a fully connected network trained with a soft-margin ranking loss on mean anomaly scores of positive and negative bags & Pixel, Flow \\ \hline
\cite{zhang2023exploiting} & Weakly-supervised & S2 & Transformer & Improves pseudo labels through completeness modeling and diversity-enhanced multi-head classification, followed by uncertainty-aware self-training that selects reliable clips using Monte Carlo Dropout & Pixel, Flow \\ \hline
TPWNG \cite{yang2024text} & Weakly-supervised & S2 & Transformer & Uses CLIP for pseudo-labeling, then trains a classifier with a Temporal Context Self-Adaptive Learning module that adjusts attention spans based on event duration & Pixel, Flow \\ \hline \hline

ESSL \cite{li2023essl} & Unsupervised & S8 & 3D CNN & Extends the jigsaw puzzle concept with a self-selective module to filter potential anomalies, enabling truly unsupervised training & Pixel \\ \hline
\cite{zaheer2022generative} & Unsupervised & S8 & MLP & A generator and discriminator co-train via cross-supervision, using pseudo-labels from reconstruction errors and a negative learning strategy to amplify anomalies & Pixel \\ \hline
\end{tabular}
\end{adjustbox}
\end{table*}

\subsection{Public Safety}
In public safety, video anomaly detection focuses on identifying risky behaviors like violence or rule violations by analyzing external cues. Pixel-based methods capture rich context but are sensitive to environment changes and privacy issues, while pose-based approaches improve robustness and privacy at the cost of visual detail. Some studies combine both in multimodal frameworks. Real-time applications demand efficient trade-offs between accuracy, privacy, and speed.

\subsubsection{\textbf{Pixel-Based Methods}}
In the context of public safety, pixel-based methods for VAD continue to play a central role due to their ability to capture fine-grained visual details, including both environmental context and object appearance. \Cref{tab:pixel_based_table} summarizes these methods. Some works pursue task-specific anomaly detection, focusing on particular threats such as shoplifting \cite{kirichenko2022detection, martinez2021criminal, muneer2023shoplifting}, weapon detection \cite{manikandan2022neural, salido2021automatic, bhatti2021weapon}, or vandalism \cite{nyajowi2021cnn}. While these methods offer high precision for well-defined scenarios, their generalizability remains limited, which motivates the mainstream research direction in VAD: detecting a broad range of anomalies without pre-defining their nature.

\begin{table*}[b]
\centering
\caption{Overview of pose-based approaches in VAD for public safety. * denotes that multiple alternative architectures have been used.}
\label{tab:pose_based_table}
\begin{adjustbox}{width=0.9\textwidth}
\begin{tabular}{>{\centering\arraybackslash}m{2.5cm}|
                >{\centering\arraybackslash}m{2.5cm}|
                >{\centering\arraybackslash}m{1cm}|
                >{\centering\arraybackslash}m{1.5cm}|
                >{\centering\arraybackslash}m{8cm}}
\hline
\textbf{Approach} & \textbf{Supervision} & \textbf{Strategy} & \textbf{Architecture} & \textbf{Distinct Characteristics / Novel Contributions} \\
\hline \hline

MoPRL \cite{yu2023regularity} & Self/Semi-supervised & S3 & Transformer & Uses a motion embedder followed by a spatio-temporal transformer for reconstruction, leveraging motion priors extracted through first-order difference statistics \\ \hline
\cite{rodrigues2020multi} & Self/Semi-supervised & S4 & CNN & Uses 1D convolutions to predict past and future poses at multiple timescales, capturing short- and long-term anomalies without relying on fixed observation windows \\ \hline
STGformer \cite{huang2022hierarchical} & Self/Semi-supervised & S4 & GCN, Transformer & Combines hierarchical spatio-temporal graphs with a two-branch architecture (local and global prediction) using spatial and temporal Transformers alongside GCNs \\ \hline
HSTGCNN \cite{zeng2021hierarchical} & Self/Semi-supervised & S4 & GCN, CNN & Uses hierarchical spatio-temporal graphs with local and global prediction branches, applying 2D temporal followed by 2D spatial graph convolutions \\ \hline
\cite{fan2021anomaly} & Self/Semi-supervised & S4 & CNN, GRU & CNN extracts spatial features, while GRU captures temporal dependencies \\ \hline
Normal Graph \cite{luo2021normal} & Self/Semi-supervised & S4 & GCN & Uses spatiotemporal graph convolution for prediction and derives anomaly scores from the prediction loss \\ \hline
PoseCVAE \cite{jain2021posecvae} & Self/Semi-supervised & S4 & CNN & Simulates anomalies via latent Gaussian mixtures, and is trained through a three-stage process combining reconstruction, KL-divergence, and binary cross-entropy losses \\ \hline
\cite{liu2021self} & Self/Semi-supervised & S6 & GCN & An autoencoder is used for feature extraction, with latent space clustering for final detection\\ \hline
STG-NF \cite{hirschorn2023normalizing} & Self/Semi-supervised & S6 & GCN & Uses normalizing flow to map inputs to a latent normal distribution, computing normality scores via likelihood and minimizing negative log-likelihood during training \\ \hline
GEPC \cite{markovitz2020graph} & Self/Semi-supervised & S6 & GCN & An autoencoder is used for feature extraction, with latent space clustering applied for VAD \\ \hline
TSGAD \cite{noghre2024exploratory} & Self/Semi-supervised & S6 & GCN & Leverages spatio-temporal graph convolution in a VAE structure, using the distance from the latent mean and variance to score anomalies based on deviation from the learned normal distribution \\ \hline
MemWGAN-GP \cite{li2023human} & Self/Semi-supervised & S3, S4 & CNN & A single-encoder dual-decoder generator with a critic, reconstructing past and predicts future sequences via memory-augmented branches \\ \hline
STGCAE-LSTM \cite{li2022human} & Self/Semi-supervised & S3, S4 & GCN, LSTM & Single-encoder, dual-decoder architecture with LSTM in the latent space for enhanced temporal analysis \\ \hline
MPED-RNN \cite{morais2019learning} & Self/Semi-supervised & S3, S4 & GRU & Uses global-local decomposition with a single encoder and dual GRU-based decoders \\ \hline
SPARTA \cite{noghre2025humancentricvideoanomalydetection} & Self/Semi-supervised & S3, S4 & Transformer & Features a single-encoder, dual-decoder transformer design and introduces a novel pose tokenization method by incorporating relative movement to emphasize motion dynamics \\ \hline
MSTA-GCN \cite{chen2023multiscale} & Self/Semi-supervised & S3, S6 & GCN & Spatio-temporal GCN and attention are used for reconstruction, with both reconstruction and latent space clustering\\
\hline
\end{tabular}
\end{adjustbox}
\end{table*}

Recent progress in weakly supervised VAD has shown that it is possible to achieve fine-grained temporal localization using only video-level labels. A prominent direction in this field involves pseudo-label refinement: Tian et al. \cite{zhang2023exploiting} propose a two-stage strategy using a multi-head classifier with diversity loss and Monte Carlo Dropout-based uncertainty filtering to generate high-quality pseudo labels. Similarly, Wang et al. \cite{shi2023abnormal} introduce ARMS, a multi-phase training framework that incrementally increases the assumed ratio of abnormal segments to progressively discover harder anomalies, supported by temporal convolution and attention. Complementary to these efforts, RTFM \cite{tian2021weakly} avoids over-reliance on classifier outputs by focusing on feature magnitudes, selecting top-k high-magnitude snippets to separate normal and abnormal segments using a multi-scale temporal architecture. In parallel, vision-language models have emerged as powerful tools for semantic alignment: Li et al. \cite{yang2024text} utilize CLIP-based feature-text alignment combined with temporal context learning, while An et al. \cite{li2022attention} adopt ViLBERT features in an MIL framework for snippet-level classification from coarse labels.

In self-supervised learning, models define proxy tasks to learn representations of normal behavior without requiring labeled anomalies, as mentioned in \Cref{sec:paradigms}. Among these, reconstruction-based methods have long been popular. To enhance reconstruction accuracy and enforce better anomaly separation, adversarial training has been widely adopted. For instance, Yang et al. \cite{yang2019enhanced} use a discriminator to distinguish between real and reconstructed patches, pushing the generator (autoencoder) to reconstruct more accurately. Chen et al. \cite{chen2021nm} instead use the discriminator to differentiate between real reconstruction error maps and synthetic noise, penalizing abnormality through structural deviations. In another approach, Georgescu et al. \cite{georgescu2021background} use irrelevant pseudo-anomalies (e.g., flowers, anime images) to train a discriminator to separate pseudo-abnormal and normal samples, encouraging the generator to focus specifically on human behavioral features.

Prediction-based models have also evolved to integrate optical flow for more accurate future frame prediction. Luo et al. \cite{luo2021future} replace basic MSE loss with a combination of flow, intensity, and gradient-based losses, alongside adversarial training for sharper predictions. Huang et al. \cite{huang2023multi} employ separate encoders for appearance and flow, feeding both into a unified decoder with skip connections and memory modules to compare current behavior with learned normal prototypes for better suppression of anomalies.

Other self-supervised tasks, such as jigsaw puzzle-based learning, aim to improve generalization by encouraging spatiotemporal reasoning. Wang et al. \cite{wang2022video} decouple spatial and temporal dimensions to form dual puzzles, solved via a 3D CNN trained to predict permutations learning both visual structure and motion patterns. Further extending generalization, some methods employ multiple proxy tasks. Georgescu et al. \cite{georgescu2021anomaly} use a suite of four self-supervised tasks: arrow-of-time prediction, motion shuffling, irregularity localization, and knowledge distillation, while its successor, SSMTL++ \cite{barbalau2023ssmtl++}, adds jigsaw puzzles and adversarial pseudo-anomalies for broader robustness. Beyond reconstruction and prediction, Doshi et al. \cite{doshi2020continual} use deep learning for feature extraction and statistical modeling to estimate the distribution of normal data, enabling adaptive decision-making through continual learning. Trained solely on normal data, the approach falls under semi-supervised learning and focuses on dynamic thresholds in evolving environments.

While these approaches reduce dependence on labeled data, even self-supervised methods often assume that training videos are purely normal. Recent research aims to relax this assumption. ESSL \cite{li2023essl} builds on puzzle-based learning but incorporates a self-selective module to identify and exclude suspected anomalies during training, enabling learning from mixed datasets. Similarly, Zaheer et al. \cite{zaheer2022generative} propose a Generative Cooperative Learning framework, where a generator reconstructs input features and a discriminator classifies them as normal or anomalous using pseudo-labels derived from reconstruction errors. A negative learning strategy intentionally trains the generator to reconstruct anomalous samples poorly, reinforcing clear distinctions between normal and abnormal patterns, achieving truly unsupervised anomaly detection. 

\subsubsection{\textbf{Pose-Based Methods}}
Pose-based VAD has emerged as a powerful alternative to appearance-based methods, particularly in applications where privacy, robustness to environmental variation, and focus on human motion are essential (see \Cref{tab:pose_based_table} for summary). The dominant paradigm in this area is semi-supervised or self-supervised learning, where models aim to learn the regular patterns of human skeletal motion using only normal data. A central challenge lies in capturing the complexity of human movement while ensuring effective generalization to unseen abnormal patterns. To this end, researchers have explored increasingly sophisticated architectures that improve reconstruction or prediction quality by modeling the temporal and spatial dynamics of the human skeleton. Given the inherent graph structure of the human pose, where joints are nodes and limbs form edges, many works \cite{noghre2024exploratory, hirschorn2023normalizing, markovitz2020graph, chen2023multiscale, huang2022hierarchical, liu2021self, zeng2021hierarchical, li2023human, li2022human, luo2021normal} naturally adopt graph-based architectures, particularly graph convolutional networks (GCNs), to model both temporal sequences and body structure.

Traditional reconstruction frameworks are extended by integrating more powerful sequence modeling mechanisms, such as transformers, which excel at capturing long-range dependencies. For instance, Yu et al. \cite{yu2023regularity} propose a tokenization scheme based on the first-order difference between pose frames and introduce a motion prior derived from training statistics to explicitly model the distribution of joint displacements, enhancing anomaly detection sensitivity.

A notable trend in recent years is the combination of reconstruction-based learning with distribution modeling. Several works \cite{chen2023multiscale, liu2021self, markovitz2020graph} adopt a two-stage framework: first, training an autoencoder on normal data and then performing latent space clustering at test time to detect anomalies as outliers. Jain et al. \cite{jain2021posecvae} utilize a variational autoencoder (VAE) to impose a probabilistic structure on the latent space, enabling more principled distribution estimation. Extending this idea further, Hirschorn et al. \cite{hirschorn2023normalizing} propose a purely probabilistic model using normalizing flows, where input pose sequences are transformed into a standard Gaussian distribution, and anomaly scores are computed via log-likelihood.

In parallel, prediction-based methods have evolved to leverage both sequential modeling and skeletal structure. Prior to the widespread adoption of GCNs, Fan et al. \cite{fan2021anomaly} used a combination of feedforward and recurrent (GRU) networks for future pose prediction. More recent works incorporate GCNs to simultaneously capture spatial (joint connectivity) and temporal (movement trajectory) patterns \cite{luo2021normal}. To further enrich the input representation, some researchers propose decomposing the pose into local (individual motion) and global (interpersonal interaction) components, as seen in \cite{huang2022hierarchical, zeng2021hierarchical}, leading to a better understanding of both individual and group behavior. Alternatively, Rodrigues et al. \cite{rodrigues2020multi} introduce a multi-timescale approach, predicting both past and future frames at varying temporal resolutions to effectively capture both short-term and long-term anomalies.

To boost overall performance, several studies adopt multi-branch architectures that combine reconstruction and prediction tasks. These systems benefit from complementary perspectives: reconstruction captures spatial structure while prediction leverages temporal dynamics. For instance, GRU-based \cite{morais2019learning}, LSTM-based \cite{li2022human}, and transformer-based \cite{noghre2025humancentricvideoanomalydetection, noghre2024posewatch} multi-branch models all report improved performance by sharing an encoder while diverging into task-specific decoders. Li et al. \cite{li2023human} enhance this design by incorporating adversarial training, aligning with trends in pixel-based VAD to improve the quality of generated sequences. Additionally, Noghre et al. \cite{noghre2024exploratory} propose a hybrid model that combines variational autoencoding for distribution-based scoring with a trajectory prediction branch, demonstrating the advantage of unifying multiple learning objectives under a coherent architecture.

Overall, pose-based VAD methods are evolving toward architectures that jointly model structure, motion, and probability, offering a privacy-aware and semantically rich alternative to pixel-level approaches. The integration of reconstruction, prediction, and distribution modeling, along with architectural innovations such as GCNs and transformer positions pose-based methods as a robust and scalable direction in VAD.
\section{Vehicle-Centric VAD}

\begin{table*}
\centering
\caption{Overview of vehicle-centric VAD approaches.}
\label{tab:vehicle}
\begin{adjustbox}{width=\textwidth}
\begin{tabular}{>{\centering\arraybackslash}m{2.5cm}|
                >{\centering\arraybackslash}m{1.5cm}|
                >{\centering\arraybackslash}m{2.5cm}|
                >{\centering\arraybackslash}m{1cm}|
                >{\centering\arraybackslash}m{2cm}|
                >{\centering\arraybackslash}m{8cm}}
\hline
\textbf{Task} & \textbf{Approach} & \textbf{Supervision} & \textbf{Strategy} & \textbf{Architecture} & \textbf{Distinct Characteristics / Novel Contributions} \\
\hline \hline

\multirow{9}{*}{\centering\arraybackslash Surveillance}
& \cite{pramanik2021real} & Supervised & S1 & - & The system detects five types of traffic anomalies (speeding, one-way violations, overtaking, illegal parking, and improper drop-offs) by combining deep learning for object detection and tracking with handcrafted algorithms \\ \cline{2-6}
& \cite{aboah2021vision} & Supervised & S1 & Decision Tree & YOLOv5 is used for object detection, while anomalies are detected using decision trees\\ \cline{2-6}
& \cite{khan2022anomaly} & Supervised & S1 & CNN & Classifies frames as accident or non-accident using a rolling average prediction algorithm \\ \cline{2-6}
& DiffTAD \cite{li2024difftad} & Self/Semi-supervised & S3 & Transformer & Models anomalies as a noisy-to-normal reconstruction process using Denoising Diffusion Probabilistic Models (DDPM), integrating Transformer-based temporal and spatial encoders to capture inter-vehicle dynamics \\ \cline{2-6}
& VegaEdge \cite{katariya2024vegaedge} & Self/Semi-supervised & S4 & GIN & A pipeline from object detection to trajectory prediction detects anomalies by comparing expected and actual trajectories \\ \cline{2-6}
& \cite{owens2000application} & Self/Semi-supervised & S6 & SOM & Trajectories are encoded into smoothed feature vectors with first and second-order motion information, allowing the SOM to detect unusual behavior by learning the distribution of normal trajectories \\ \cline{2-6}
& \cite{hu2006system} & Unsupervised & S7 & - & Uses DL methods for feature extraction and later uses a hierarchical clustering method based on K-means, modeling each motion pattern as a chain of Gaussian distributions, and enabling both anomaly detection and behavior prediction \\ \cline{2-6}
& \cite{wang2007Unsupervised} & Unsupervised & S8 & Bayesian Model & A hierarchical Bayesian framework that uses LDA and HDP to jointly model atomic activities and multi-agent interactions without requiring labeled data \\ \cline{2-6}
& \cite{santhosh2021vehicular} & Self/Semi-supervised & S3, S6 & CNN & Converts vehicle trajectories into gradient images, leverages a CNN to classify normal trajectories via unsupervised clustering, and uses a VAE to detect unseen anomalies through reconstruction loss \\

\hline \hline
\multirow{8}{*}{\centering\arraybackslash Autonomous Driving}
& \cite{zhou2022spatio} & Supervised & S1 & CNN, MLP, SVM & Temporal features are clustered after MLP processing to identify potential accidents, which are then combined with CNN-based spatial features and classified using an SVM \\ \cline{2-6}
& \cite{park2024deep} & Supervised & S1 & CNN & Combines YOLOv5, lane alignment, and motion tracking to detect stopped vehicles \\ \cline{2-6}
& TempoLearn \cite{htun2023tempolearn} & Supervised & S1 & CNN, LSTM, Transformer & Uses CNN and LSTM for spatiotemporal feature extraction and a Transformer classifier for accident detection \\ \cline{2-6}
& \cite{fang2022traffic} & Self/Semi-supervised & S4 & CNN, LSTM & Collaborative multi-task framework for jointly predicting future frames, object locations, and scene context \\ \cline{2-6}
& FOL \cite{yao2019unsupervised} & Self/Semi-supervised & S4 & CNN & The future expected trajectory is compared to the actual trajectory for detecting abnormal behaviors \\ \cline{2-6}
& \cite{ru2024enhanced} & Self/Semi-supervised & S4 & Transformer & A dual GAN framework with a Swin-Unet-based generator to predict intermediate frames using both optical flow and cropped inputs \\ \cline{2-6}
& $\mathrm{HF}^{2}\text{-}\mathrm{VAD}$ \cite{bogdoll2024hybrid} & Self/Semi-supervised & S3, S4 & CNN & Combines memory-augmented autoencoders for reconstruction and conditional VAEs for future frame prediction, enabling fine-grained, dense anomaly localization \\ \cline{2-6}
& \cite{haresh2020towards} & Self/Semi-supervised & S3, S6 & 3D CNN, GCN & Proposes two models: one based on manifold learning to identify out-of-distribution anomalies, and another using reconstruction to detect deviations from normal data \\
\hline
\end{tabular}
\end{adjustbox}
\end{table*}

\begin{table*}[]
\centering
\caption{Overview of reviewed work in fire and flood detection. Architectures are inferred from reported methodologies when possible.}
\label{tab:fire_flood_detection}
\begin{adjustbox}{width=0.9\textwidth}
\begin{tabular}{>{\centering\arraybackslash}m{2cm}|
                >{\centering\arraybackslash}m{2cm}|
                >{\centering\arraybackslash}m{1.5cm}|
                >{\centering\arraybackslash}m{10cm}}
\hline
\textbf{Task} & \textbf{Approach} & \textbf{Architecture} & \textbf{Distinct Characteristics / Novel Contributions} \\
\hline\hline
\multirow{17}{*}{Fire Detection}
& \cite{khalil2021fire} & GMM & Uses GMM motion detection and a region growth tracking, enabling accurate fire segmentation, growth rate estimation, and significant false alarm reduction \\ \cline{2-4}
& \cite{huang2022fire} & CNN & Integrates 2D Haar wavelet transforms with convolutional neural networks to combine spatial and spectral features, achieving higher detection accuracy and significantly reducing false alarms and computational complexity \\ \cline{2-4}
& Fire-Det \cite{gao2024two} & CNN & A two-stage framework combining motion detection with specialized Fire-Det and lightweight Fire-Det Nano models, enabling fast, accurate early fire detection\\ \cline{2-4}
& E-EFNet \cite{farman2023efficient} & CNN & Based on EfficientNetB0 enhanced with stacked autoencoders and dense connections, achieving high accuracy, reduced false alarms, and efficient real-time inferencing \\ \cline{2-4}
& \cite{chitram2024enhancing} & CNN & Combining EfficientNet and YOLOv5, leveraging compound scaling and real-time object detection \\ \cline{2-4}
& \cite{mahdi2022edge} & CNN & A modified YOLOv5 model within an edge computing framework using Jetson Nano, featuring a dropout-enhanced architecture for improved accuracy and speed, and integration with cloud services for real-time alerting \\ \cline{2-4}
& \cite{dou2024improved} & CNN & An improved YOLOv5s-based fire detection model that enhances detection accuracy and efficiency by integrating CBAM, BiFPN, and transposed convolution \\ \cline{2-4}
& \cite{sathishkumar2023forest} & CNN & Proposes Learning Without Forgetting (LWF) framework to enable transfer learning \\ \cline{2-4}
& \cite{son2018video} & CNN & Analyzes the efficacy of famous networks such as AlexNet, GoogLeNet, and VGG-16 \\ \cline{2-4}
& FSDF \cite{zhao2024fsdf} & CNN & Fuses traditional HSV and CLBP-based feature enhancement with deep learning models YOLOv8 and VQ-VAE, achieving high precision and robustness \\ \cline{2-4}
& \cite{yunusov2024robust} & CNN & Combines transfer learning with YOLOv8 and the TranSDet model, incorporating a boosting-based ensemble to enhance small and large fire detection accuracy \\ \cline{2-4}
& \cite{akhmedov2024dehazing} & CNN & Integrates a dehazing algorithm with a fine-tuned YOLO-v10 for ship fire detection \\ \cline{2-4}
& \cite{yar2024efficient} & 3D CNN & A modified MobileNetV3 integrated with a 3D CNN and a novel soft attention mechanism, enhancing spatial awareness and reducing model complexity \\ \cline{2-4}
& FWSRNet \cite{wang2024computer} & Transformer & A Vision Transformer-based model incorporating self-attention and contrastive feature learning for fine-grained wildfire and smoke recognition \\ \cline{2-4}
& \cite{yar2024modified} & Transformer & A modified vision transformer architecture for fire detection that enables learning from scratch on small to medium-sized datasets by integrating shifted patch tokenisation and locality self-attention \\
\hline\hline
\multirow{8}{*}{Flood Detection}
& \cite{filonenko2015real} & - & Combines background subtraction, morphological operations, color probability modeling across, and spatial features like edge density and boundary roughness \\ \cline{2-4}
& \cite{borges2008probabilistic} & Bayesian Classifier & A probabilistic model for flood detection by combining spatial features with temporal variation and a non-central chi-square-based positional prior, using Bayes classification and patch-level scoring \\ \cline{2-4}
& V-FloodNet \cite{liang2023v} & CNN & A video segmentation system that uses template-matching-based water depth estimation method, enabling accurate urban flood detection \\ \cline{2-4}
& FRAD \cite{villalon2021convolutional} & CNN & Applies a CNN network to high-resolution multispectral remote sensing images (SPOT-5) for supervised classification of urban flood risk \\ \cline{2-4}
& \cite{zhong2024detection} & CNN & A YOLOv4-based deep learning method for urban flood depth estimation using traffic images, leveraging submerged reference objects \\ \cline{2-4}
& \cite{lohumi2018automatic} & CNN, GRU & Flood severity classification from videos, combining spatial features extracted by a modified VGG16-based CNN with temporal dependencies captured by GRUs \\ \cline{2-4}
& \cite{lopez2017multi} & CNN, LSTM & Flood detection in social media by combining visual features using a fine-tuned InceptionV3 CNN with semantic features from metadata using a bidirectional LSTM \\
\hline
\end{tabular}
\end{adjustbox}
\end{table*}

\subsection{Road Surveillance}
In the context of vehicle VAD for road surveillance, systems are primarily utilized by traffic monitoring authorities and urban infrastructure. These systems demand high-resolution spatial coverage, real-time or near-real-time processing, and robustness under diverse environmental conditions. The corresponding responses are typically passive and retrospective, including alert generation, traffic violation reporting, or data archiving for forensic purposes.

Early methods in vehicle-focused VAD followed similar trajectories to general VAD research, relying on handcrafted features and rule-based logic (see \Cref{tab:vehicle}). These approaches often combine object detection and tracking with non-deep learning models for classification. Pramanik et al. \cite{pramanik2021real} employed five distinct algorithms to identify specific behaviors such as speed violations and illegal parking. Zhou et al. \cite{zhou2022spatio} utilized Support Vector Machines (SVMs) for accident detection based on extracted spatial-temporal features. Although effective for narrowly defined tasks, these approaches lack flexibility and generalization to unforeseen behaviors. While early deep learning-based models such as \cite{khan2022anomaly} extended detection capabilities, they still largely operated within constrained anomaly categories.

To mitigate these limitations, Aboah et al. \cite{aboah2021vision} proposed a decision-tree-based method that evaluates foreground and background object detections using spatial thresholds and Intersection-over-Union (IoU) metrics, offering a more adaptable and interpretable rule-based framework.

A more flexible perspective involves casting anomaly detection as an unsupervised clustering problem, where anomalies are treated as statistical outliers. These methods utilize features such as motion trajectories, foreground activity, or background dynamics to learn normal patterns without requiring explicit labels. Hu et al. \cite{hu2006system} applied k-means clustering to vehicle trajectories, while Niebles et al. \cite{wang2007Unsupervised} adopted hierarchical Bayesian models—originally developed for language modeling—to cluster interactions and motions, thereby learning the structure of normal behavior in a probabilistic manner. Such clustering-based methods offer greater adaptability in complex or evolving environments.

More recent work has shifted toward deep learning models that emphasize generalization. Many of these methods fall into the prediction/reconstruction-based paradigms discussed in \Cref{sec:paradigms}. Santhosh et al. \cite{santhosh2021vehicular} employed a variational autoencoder to reconstruct trajectory data, while Li et al. \cite{li2024difftad} leveraged diffusion models to learn the data distribution. In the prediction domain, Katariya et al. \cite{katariya2024vegaedge} used a graph isomorphism network with attention mechanisms to model interactions and forecast future trajectories, and Fang et al. \cite{fang2022traffic} proposed a multi-task learning framework that predicts future frames, object locations, and scene context simultaneously.


\subsection{Autonomous Driving}
In autonomous driving, Vehicle VAD serves as a real-time, safety-critical component integrated into the vehicle's decision-making pipeline. These systems demand low-latency processing, precise detection of complex motion patterns, and seamless integration with multi-sensor fusion modules including LiDAR, radar, and cameras.

Early research in this domain primarily focused on detecting well-defined types of anomalies, often formulated as supervised classification (see \Cref{tab:vehicle}). Park et al. \cite{park2024deep} addressed the detection of stopped vehicles using dense optical flow to estimate host vehicle motion and bounding-box analysis to track surrounding vehicles. Similarly, some works have narrowed their scope to the detection and categorization of different types of collisions. Htun et al. \cite{htun2023tempolearn} proposed a deep learning architecture that uses CNNs and LSTMs to extract spatial and temporal features, respectively, followed by a region proposal module and a classification head to detect and categorize collision types.

Building upon these constrained approaches, more flexible systems have emerged. Zhou et al. \cite{zhou2022spatio} introduced a two-stage coarse-to-fine framework: the first stage performs clustering of encoded temporal features to identify outlier frames as potential anomaly candidates, while the second stage applies object-level spatial feature extraction and a trained SVM classifier to confirm accident frames.

Reconstruction-based methods have also gained traction due to their generalization capacity to unseen anomalies. Haresh et al. \cite{haresh2020towards} enhanced traditional autoencoder architectures by incorporating region proposal networks for object detection and graph convolutional networks (GCNs) to model object interactions, improving the semantic richness of reconstructions.

One of the most adopted modalities is optical flow, which provides dense motion information. Optical flow enables the detection of sudden or abnormal motion patterns, making it useful across both prediction and reconstruction-based paradigms. Bogdoll et al. \cite{bogdoll2024hybrid} proposed a convolutional variational autoencoder that fuses features from both RGB and optical flow, improving anomaly reconstruction. In prediction-based frameworks, Yao et al. \cite{yao2019unsupervised} leveraged optical flow for future object localization and ego-motion prediction, detecting anomalies based on deviations from expected motion trajectories. Ru et al. \cite{ru2024enhanced} extended this idea through a dual-GAN framework that jointly predicts both optical flow and appearance features in regions of interest using a Swin-Unet backbone, achieving high accuracy at the cost of computational efficiency. Similarly, Fang et al. \cite{fang2022traffic} proposed a multi-task framework incorporating future frame prediction, motion trajectory consistency, and visual context modeling, with optical flow as a core feature to enhance anomaly detection performance.

Despite the demonstrated success of prediction-based models, their performance can degrade in highly dynamic environments involving complex multi-agent interactions or unexpected environmental changes. These limitations are particularly pronounced in ego-centric settings, where the camera is mounted on a moving vehicle, increasing the risk of false positives due to background motion or occlusions.

\begin{figure*}[]
\centering
\includegraphics[clip,trim={18 20 5 5},width=0.9\textwidth]{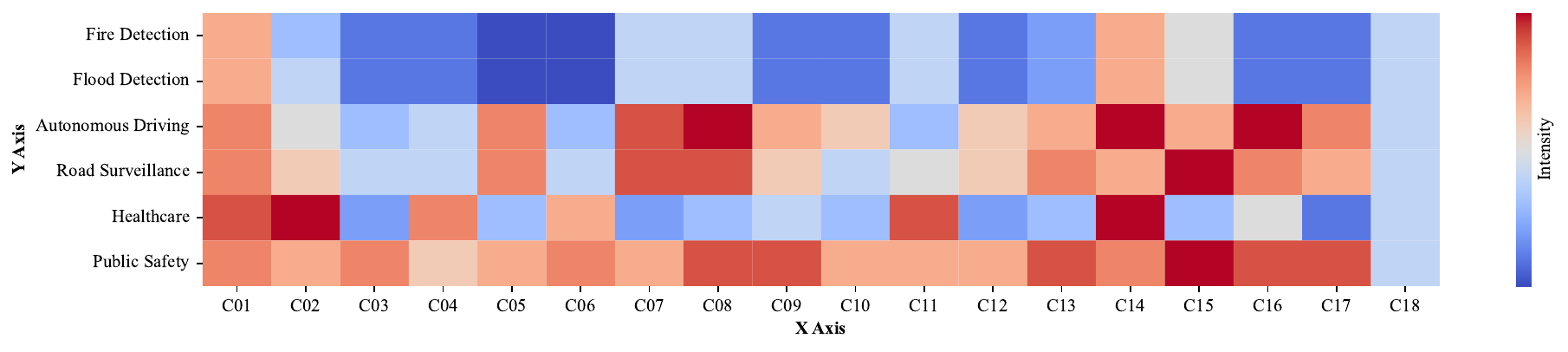}
\caption{Severity of Each Challenge Across Different VAD Domains.}
\label{fig:heatmap}
\end{figure*}
\section{Environmental-Centeric VAD}


Environmental VAD is critical for enabling rapid response and minimizing harm, particularly during the real-time detection stage of disaster management. Unlike prediction and post-event assessment, which rely on early indicators or support recovery and fall outside the scope of this work, real-time detection has been primarily approached through supervised methods that treat specific disasters as classification tasks. Video-based fire and flood detection have received the most attention in computer vision due to their structured visual signatures, whereas disasters like hurricanes and tsunamis are better suited to satellite imagery, and events like earthquakes and droughts often lack distinct visual cues. This survey focuses on video-based methods for detecting floods and fires (see \Cref{tab:fire_flood_detection}), where vision remains a central and effective modality.

\subsection{Fire Detection}

CNN-based methods have been explored for fire detection due to their ability to capture spatial features. Early works utilized pretrained CNN models such as MobileNetV2 \cite{sandler2018mobilenetv2}, AlexNet \cite{krizhevsky2012imagenet}, and GoogLeNet \cite{szegedy2015going}, fine-tuning them for fire detection tasks. Some studies combined these models with additional techniques to enhance performance. For instance, wavelet transforms were used to extract critical spectral features \cite{huang2022fire}, while transfer learning with "learning without forgetting" ensured models retained prior knowledge when adapting to new environments \cite{son2018video}. Advanced approaches integrated 3D CNNs with modified attention mechanisms to improve accuracy and employed Grad-CAM for visual interpretability of model decisions \cite{yar2024efficient}.

Another group of studies \cite{chitram2024enhancing, mahdi2022edge, yunusov2024robust, akhmedov2024dehazing} defined fire detection as a subset of object detection, leveraging and adapting well-known algorithms such as Faster R-CNN \cite{ren2016faster} and YOLO \cite{redmon2016you} for this purpose. For instance, \cite{dou2024improved} extended YOLOv5s by incorporating Convolutional Block Attention Modules (CBAM) \cite{woo2018cbam} for improving feature fusion and replacing nearest neighbor interpolation with transposed convolution, introducing a fast, compact model and a more complex, accurate version tailored for fire detection. Similarly, \cite{zhao2024fsdf} utilized YOLOv8 as a feature extractor to identify regions likely to contain fire. While this approach alone can serve as a fire detection method, they augmented it with a Vector Quantized Autoencoder (VQ-VAE) \cite{van2017neural} to model the distribution of fire patterns, thereby providing an additional layer of analysis to reduce false positives and enhance detection reliability.

With the growing popularity of transformers, several studies have begun employing the Vision Transformers (ViT) \cite{dosovitskiy2020image} for fire detection. \cite{yar2024modified} introduces a specialized tokenization method designed for effective input tokenization for transformers. Additionally, \cite{wang2024computer} utilizes a contrastive feature learning mechanism to enhance the model's discriminative capabilities.

Earlier works treated video as independent frames, ignoring temporal dynamics. Recent studies, however, emphasize the importance of motion. For example, \cite{khalil2021fire} uses segmentation and GMMs to identify flame-like motion and estimate fire growth, while \cite{gao2024two} applies GMMs for motion filtering before fire classification. These approaches demonstrate how incorporating temporal cues enhances accuracy and context awareness.

\subsection{Flood Detection}

Earlier works \cite{filonenko2015real, borges2008probabilistic} rely on fundamental probabilistic models and heuristic approaches. These methods focus on extracting visual features like color, texture, and motion to identify flood regions. \cite{borges2008probabilistic} integrates color, texture, and dynamic features within a probabilistic framework, leveraging spatial distributions to enhance detection accuracy. \cite{filonenko2015real} employs background subtraction, morphological processing, and boundary roughness analysis for improved efficacy. 

Deep learning brought strong advancements \cite{lohumi2018automatic, villalon2021convolutional, lopez2017multi}. \cite{lohumi2018automatic} utilizes a hybrid CNN-GRU model to classify flood severity in videos, combining spatial feature extraction with temporal modeling of sequential frames. \cite{villalon2021convolutional} adopts a CNN-based Flood-Risk Assessment and Detection (FRAD) method for processing multispectral satellite images to identify flood-risk zones, emphasizing urban planning applications. \cite{lopez2017multi} combines visual CNN-based analysis with a BiLSTM network for textual metadata processing, creating a multimodal approach to flood detection. These approaches exemplify the power of deep learning in capturing both spatial and temporal intricacies, demonstrating significant improvements over traditional models in accuracy and versatility.

More advanced architectures \cite{liang2023v, humaira2023dx, zhong2024detection}, incorporate SOTA techniques to enhance flood detection capabilities. \cite{liang2023v} proposes V-FloodNet, a system integrating video segmentation (AFB-URR) and image segmentation (EfficientNet-B4 and LinkNet) with novel template-matching for depth estimation. \cite{humaira2023dx} introduces DX-FloodLine, combining VGG16-LSTM for flood classification and Faster R-CNN with Mask R-CNN for object detection. \cite{zhong2024detection} applies YOLOv4 for urban flood detection, utilizing traffic images with submerged reference objects and achieving real-time performance.

\section{Conclusions and Future Directions}
This survey provides a comprehensive and structured overview of deep learning-based Video Anomaly Detection (VAD), examining major challenges, learning paradigms, and a range of application domains. By incorporating human-, vehicle-, and environment-centric perspectives, it reveals both shared foundations and domain-specific characteristics, facilitating meaningful cross-domain insights. The proposed taxonomy of supervision levels and adaptive strategies clarifies the strengths and limitations of existing methods, offering actionable guidance for designing effective VAD systems. In identifying critical research gaps, this work outlines promising directions for future exploration and serves as both a primer for newcomers and a valuable reference for researchers seeking to build robust, scalable solutions for real-world applications. Building on the challenges and trends observed across different VAD domains, we further evaluate the severity of open problems, as visualized in \Cref{fig:heatmap}, to support strategic research planning. Environment-centric VAD tends to be more manageable due to its structured, constrained settings. In contrast, autonomous driving remains highly challenging due to issues like domain shift, real-time performance, and sensor calibration (C8, C14, C16). Large-scale deployment in road surveillance and public safety introduces major scalability concerns (C15), driving the development of resource-efficient models, along with alternative data modalities. In healthcare, annotation remains a significant bottleneck (C2) due to the dependence on expert knowledge, underscoring the importance of label-efficient approaches such as few-shot and weakly supervised learning. Moreover, data scarcity (C1) persists across nearly all domains, prompting increased interest in synthetic data generation, especially with generative AI to simulate anomalies and boost model robustness.

\section*{Acknowledgments}
This research is funded by the United States National Science Foundation (NSF) under award number 2329816.

 
\bibliographystyle{IEEEtran}
\bibliography{IEEEabrv,ref}

\begin{IEEEbiography}[{\includegraphics[width=1in,height=1.25in,keepaspectratio]{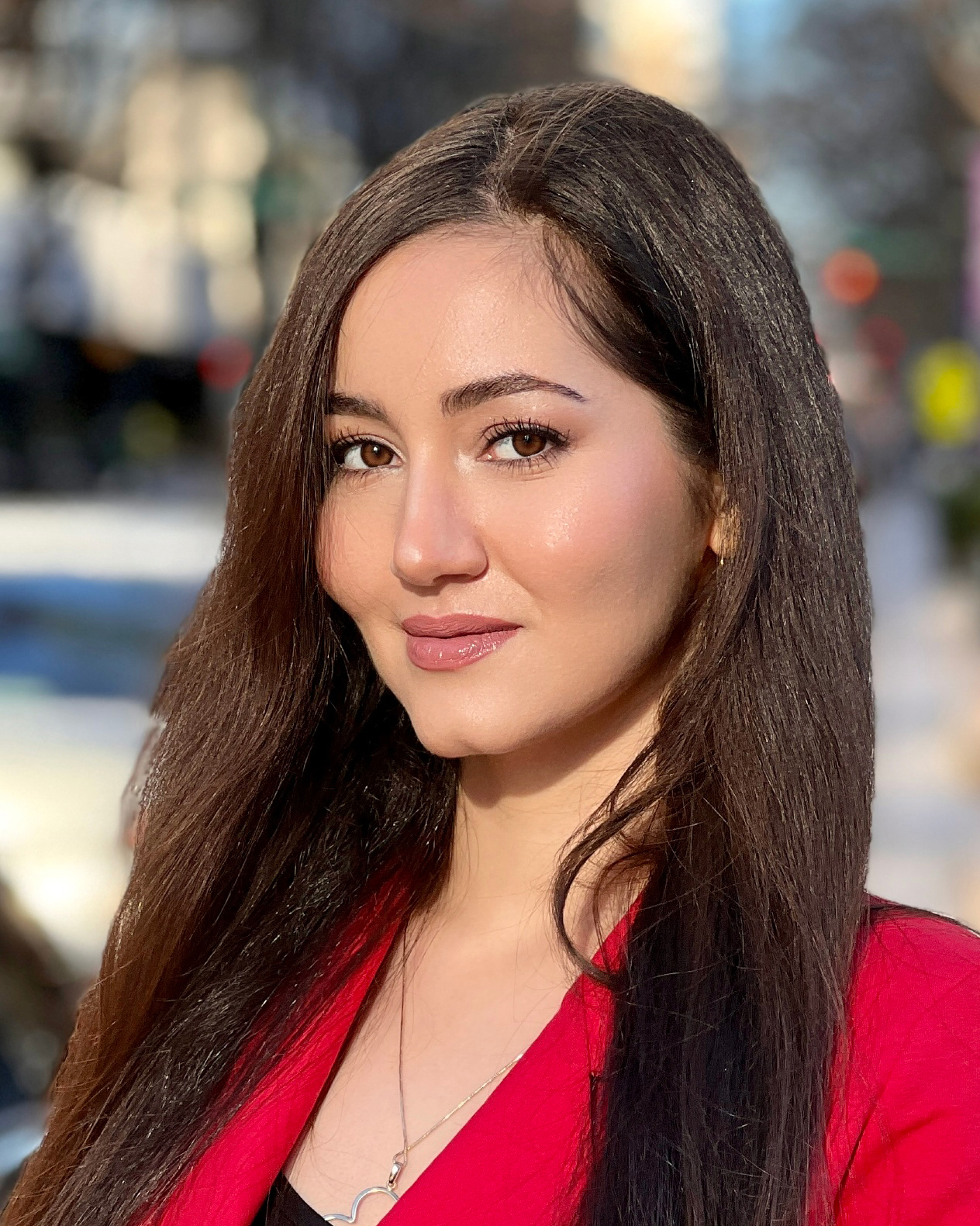}}]{Ghazal Alinezhad Noghre} is currently a Ph.D. candidate in Electrical and Computer Engineering at the University of North Carolina at Charlotte, Charlotte, North Carolina, United States. Her research focuses on artificial intelligence, machine learning, and computer vision, with a particular emphasis on the application of AI in real-world environments and the associated challenges.
\end{IEEEbiography}

\vspace{-40pt}
\begin{IEEEbiography}[{\includegraphics[width=1in,height=1.25in,keepaspectratio]{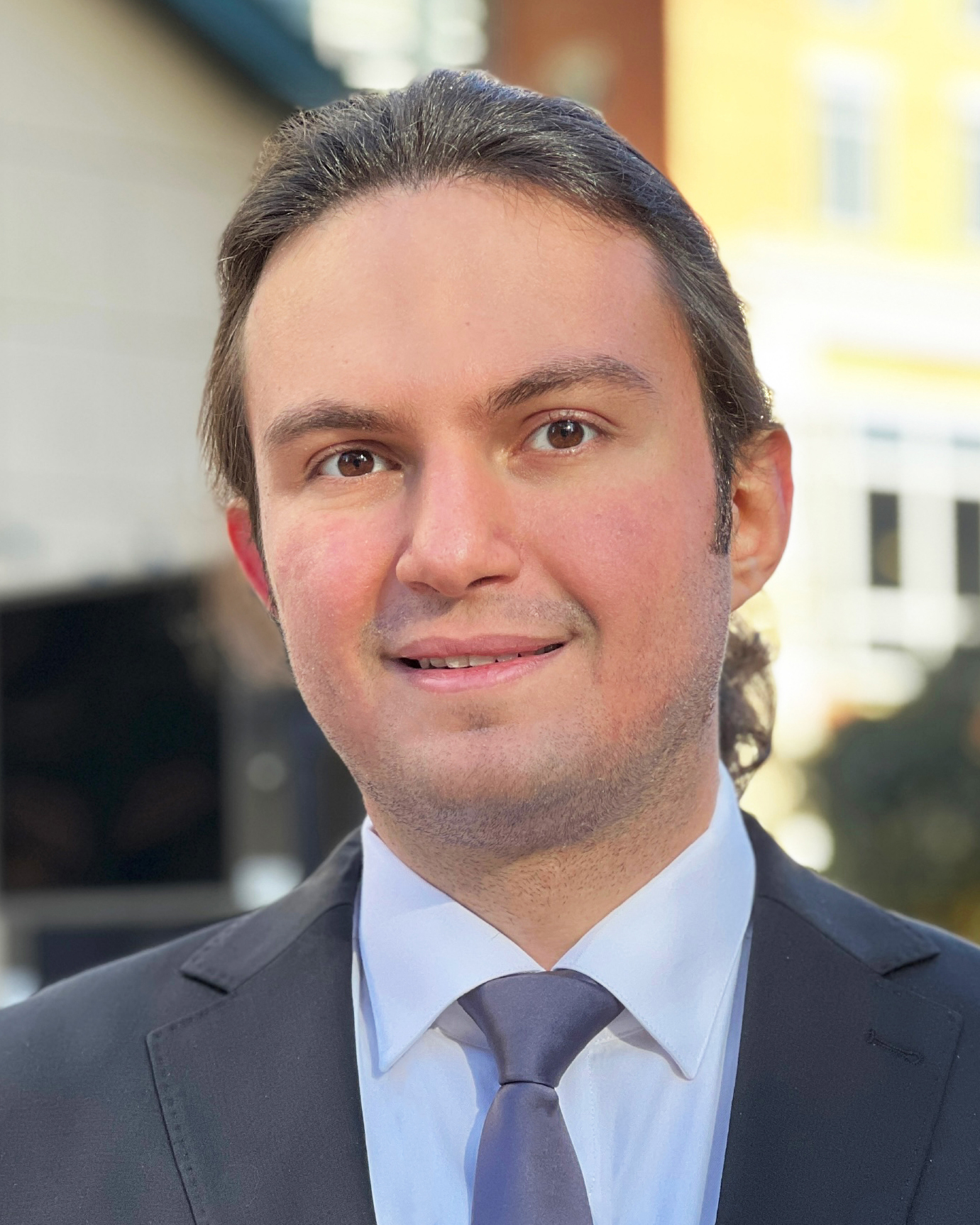}}]{Armin Danesh Pazho} is a Ph.D. candidate in Electrical and Computer Engineering at the University of North Carolina at Charlotte. His research focuses on artificial intelligence, machine learning, and computer vision, with emphasis on developing scalable AI solutions for practical applications. He has researched, designed, and developed novel AI/ML algorithms, systems, and datasets with deployment in real-world testbeds.
\end{IEEEbiography}

\vspace{-40pt}
\begin{IEEEbiography}[{\includegraphics[width=1in,height=1.25in,keepaspectratio]{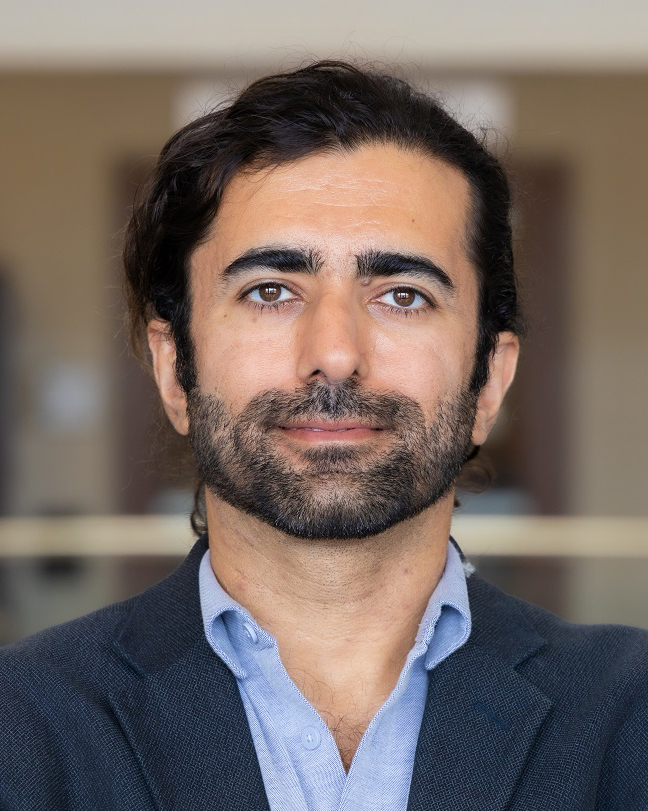}}]{Hamed Tabkhi} is an associate professor of Electrical and Computer Engineering at the University of North Carolina at Charlotte. His research focuses on advancing artificial intelligence and computer vision to solve real-world challenges through close collaboration with experts and community stakeholders. The National Science Foundation recognized Dr. Tabkhi's Smart and Connected Communities award as a program success story. His work has been featured by local news for its significant contributions to community-driven responsible AI solutions.
\end{IEEEbiography}

\newpage

 




\vfill







\title{A Survey on Video Anomaly Detection via Deep Learning: Human, Vehicle, and Environment: Supplementary Materials}

\author{Ghazal Alinezhad Noghre$^1$, Armin Danesh Pazho$^1$, Hamed Tabkhi$^1$
\thanks{$^1$ Electrical and Computer Engineering Department, UNC Charlotte (galinezh, adaneshp, htabkhiv@charlotte.edu)}
}

\markboth{IEEE Transactions On Pattern Analysis and Machine Intelligence}%
{Noghre \MakeLowercase{\textit{et al.}}: VAD via Deep Learning Survey}


\maketitle

\section{Abbreviations}

This section provides a list of abbreviations and their corresponding full forms used throughout the survey. These terms are commonly referenced in the literature on Video Anomaly Detection (VAD). The purpose of this list is to assist readers with quick reference and improve the clarity and accessibility of the material presented.

\begin{table*}[]
\centering
\caption{List of abbreviations used throughout the paper. This table provides full forms for technical terms commonly referenced in the context of Video Anomaly Detection (VAD).}
\label{tab:abbreviations_table}
\begin{adjustbox}{width=0.9\textwidth}
\begin{tabular}{>{\centering\arraybackslash}m{3cm}|>{\arraybackslash}m{13cm}}
\hline
\textbf{Abbreviation} & \textbf{Full Form} \\
\hline \hline
VAD & Video Anomaly Detection \\
SOTA & State-of-The-Art \\
AI & Artificial Intelligence \\
CNN & Convolutional Neural Network \\
VQ-VAE & Vector Quantized Variational Autoencoder \\
ViT & Vision Transformer \\
GMM & Gaussian Mixture Model \\
GRU & Gated Recurrent Unit \\
LSTM & Long Short-Term Memory \\
RGB & Red Green Blue (color video input) \\
MLP & Multi-Layer Perceptron \\
PD & Parkinson’s Disease \\
SVM & Support Vector Machine \\
GEI & Gait Energy Image \\
ASD & Autism Spectrum Disorder \\
EEG & Electroencephalography \\
VEEG & Video Electroencephalography \\
MIL & Multiple Instance Learning \\
MSE & Mean Squared Error \\
GCN & Graph Convolutional Network \\
VAE & Variational Autoencoder \\
GIN & Graph Isomorphism Network \\
kNN & k-Nearest Neighbors \\
kDNN & k-Nearest Distance Neural Network (DNN-based approximation of kNN) \\
IoU & Intersection-over-Union \\
GAN & Generative Adversarial Network \\
\hline
\end{tabular}
\end{adjustbox}
\end{table*}


\end{document}